\documentclass{article}

\usepackage{iclr2027/iclr2027_conference,times}

\usepackage[utf8]{inputenc}
\usepackage[T1]{fontenc}
\usepackage{microtype}
\usepackage{placeins}
\usepackage{wrapfig}
\usepackage{needspace}
\usepackage{graphicx}
\usepackage{booktabs}
\usepackage{multirow}
\usepackage{url}
\usepackage[table]{xcolor}
\usepackage{hyperref}
\hypersetup{hidelinks}

\definecolor{gain}{rgb}{0.8,0.0,0.0}
\definecolor{loss}{rgb}{0.0,0.6,0.0}
\newcommand{\dg}[1]{\,\textcolor{gain}{\scriptsize(#1)}}
\newcommand{\dl}[1]{\,\textcolor{loss}{\scriptsize(#1)}}

\usepackage{listings}
\usepackage{amsmath}
\usepackage{amsfonts}
\usepackage{amssymb}
\usepackage{nicefrac}

\usepackage{algorithm}
\usepackage{algpseudocode}

\usepackage{tabularx}
\usepackage{afterpage}
\usepackage[most]{tcolorbox}
\newtcolorbox{casetrace}[1]{%
  enhanced,
  colback=white,
  colframe=black!25,
  colbacktitle=black!5,
  coltitle=black,
  title={#1},
  fonttitle=\small\bfseries,
  fontupper=\small,
  boxrule=0.45pt,
  arc=1pt,
  left=8pt, right=8pt, top=7pt, bottom=7pt,
  toptitle=4pt, bottomtitle=4pt,
  boxsep=0pt,
  before skip=10pt, after skip=10pt,
}

\title{Beyond Math and Code: Lightweight\\ Corpus-Grounded Process Rewards\\ for Factual Question Answering}

\author{%
  \begin{minipage}[t]{\dimexpr\textwidth-2\tabcolsep\relax}
    \centering\normalfont
    {\normalsize\bfseries
      Shicheng Fan$^{1*\dagger}$ \quad Haochang Hao$^{1*}$ \quad
      Dehai Min$^{2,1*}$ \quad Weihao Liu$^1$ \\[0.35ex]
      Hanrong Zhang$^{3,1}$ \quad Lingwei Wei$^4$ \quad
      Henry Peng Zou$^1$ \quad Chengquan Guo$^5$ \\[0.35ex]
      Jie Yang$^1$ \quad Honghui Bao$^1$ \quad Zhiwei Liu$^6$ \quad
      Lu Cheng$^7$ \quad Philip S.\ Yu$^{1\dagger}$\endgraf
    }
    \vspace{1.1ex}
    {\small
      $^1$University of Illinois at Chicago \quad $^2$ByteDance Inc. \\
      $^3$Google \quad $^4$Institute of Information Engineering \\
      $^5$University of Chicago \quad $^6$Recursive Superintelligence \\
      $^7$Pennsylvania State University\endgraf
    }
    \vspace{0.8ex}
    {\footnotesize
      \texttt{\{sfan25,hhao,dmin10,wliu681,hzhan135\}@uic.edu} \\
      \texttt{\{pzou3,jyang265,hbao,psyu\}@uic.edu} \\
      \texttt{weilingwei@iie.ac.cn} \quad \texttt{chengquanguo@uchicago.edu} \\
      \texttt{zhiweiliu@recursive.com} \quad \texttt{lqc5822@psu.edu}\endgraf
    }
    \vspace{0.8ex}
    {\footnotesize
      $^*$Equal contribution. \quad
      $^\dagger$Correspondence: \texttt{\{sfan25,psyu\}@uic.edu}\endgraf
    }
  \end{minipage}%
}

\iclrfinalcopy 

\begin{document}
\maketitle
\lhead{Preprint}


\begin{abstract}
Process supervision during reinforcement learning (RL) post-training matters in factual question answering (QA) because responses receiving positive response-level rewards can still contain sentence-level factual errors. Unlike math and code, factual QA lacks inexpensive programmatic checks, making reward computation a bottleneck. Existing methods rely on neural verifiers to score individual sentences, requiring extensive model inference as RL repeats these checks across many sampled responses at every update. We therefore introduce \textbf{CorVer}\footnote{Code: \url{https://github.com/shichengf/CorVer}.} (\emph{Corpus Verify}), a lightweight training-time approach to process supervision that derives sentence-level rewards from corpus co-occurrence statistics. A 0.5B extractor identifies subject--object pairs, indexed corpus queries supply their co-occurrence counts, and the resulting sentence rewards are assigned to the corresponding tokens for RL. On Qwen3-4B and Qwen3-8B, CorVer reduces mean complete training time by $5.5$--$10.4\times$ relative to the four factuality-RL baselines. Across the four models evaluated against these baselines, CorVer achieves the highest accuracy in $17$ of $20$ model--benchmark settings. CorVer outperforms the unmodified models in all $30$ standard factual-QA settings (six models from three families across five benchmarks) and remains effective on two additional multi-hop QA datasets.

\end{abstract}
\section{Introduction}
\label{sec:intro}

Reinforcement learning (RL) post-training of large language models (LLMs) often rewards correct final answers~\citep{deepseek-ai_deepseek-r1_2025}, even when the reasoning is flawed. Incorporating process supervision into RL provides rewards for intermediate steps~\citep{lightman_lets_2023}, but repeated verification can make reward computation a training bottleneck. For example, $100$ updates with $3$ questions, $16$ responses per question, and an assumed five sentences per response entail $24{,}000$ verifier evaluations. Programmatic checks can make verification inexpensive in math and code, but factual question answering (QA) lacks such simple checks for the factual correctness of individual sentences (Figure~\ref{fig:intro-concept}). Existing methods commonly rely on neural verifiers, such as LLM-as-judge models, to compute process rewards~\citep{wu2025comprehensive}. Compared with these programmatic checks, LLM-based verification requires substantial inference time for each sentence, making repeated evaluation across sampled responses and RL updates expensive. Moreover, the response generator and the judge are both LLMs, and similar pretraining data may leave them with shared knowledge gaps. The judge may therefore fail to detect factual errors arising from these gaps.

\begingroup
\setlength{\intextsep}{6pt}
\begin{figure}[H]
\begin{minipage}[t]{0.58\textwidth}
  \vspace{0pt}
These limitations motivate exploring cheaper supervision signals. Prior work links corpus co-occurrence to factual recall~\citep{kang2023impact} and uses it for inference-time retrieval~\citep{min2025quco}. In a manual audit of $700$ sentences from QA responses, factual correctness rises monotonically from $24\%$ at zero subject--object co-occurrences to $81\%$ at counts of at least $20$ (Figure~\ref{fig:calibration-curve}). These counts provide external information for sentence-level rewards without relying on a judge's factual recall.
\end{minipage}\hfill
\begin{minipage}[t]{0.38\textwidth}
  \vspace{0pt}
  \centering
  \setlength{\parskip}{0pt}
  \includegraphics[width=\linewidth]{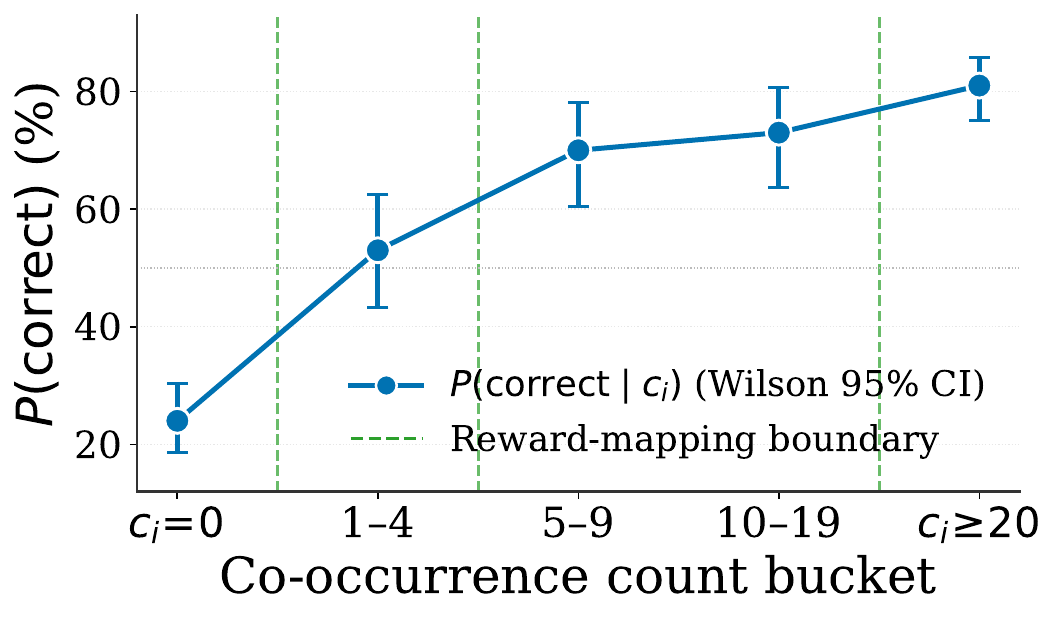}\par
  \nointerlineskip
  \setlength{\abovecaptionskip}{4pt}
  \caption{\textbf{Counts track factuality.}}
  \label{fig:calibration-curve}
\end{minipage}
\end{figure}
\endgroup

\newlength{\introPanelHeight}
\begin{figure}[t]
  \settoheight{\introPanelHeight}{\includegraphics[width=0.55\textwidth]{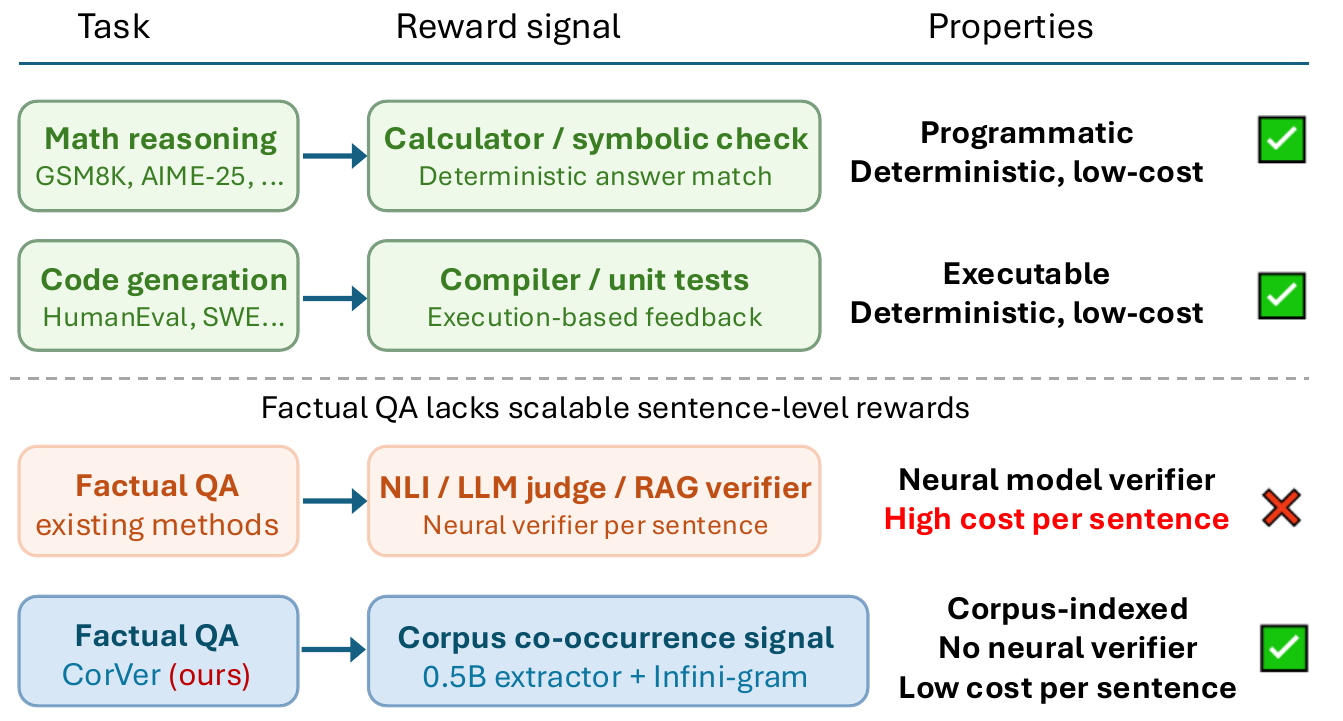}}
  \centering
  \begin{minipage}[b]{0.55\textwidth}
    \centering
    \includegraphics[width=\linewidth]{figure/figure_1.pdf}
    \par\smallskip
    {\small (a) Reward design}
  \end{minipage}\hfill
  \begin{minipage}[b]{0.43\textwidth}
    \centering
    \includegraphics[height=\introPanelHeight,width=\linewidth,keepaspectratio]{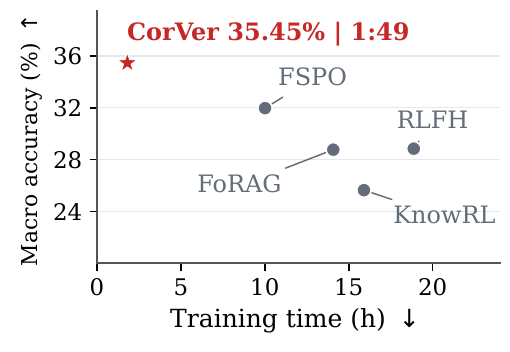}
    \par\smallskip
    {\small (b) Accuracy and training cost}
  \end{minipage}
  \caption{\textbf{Corpus rewards for accurate and efficient factual QA.} (a) Math and code admit inexpensive programmatic checks, whereas existing factual-QA methods rely on neural verification for sentence-level rewards. CorVer instead uses a corpus-indexed co-occurrence signal without a neural verifier. NLI: natural language inference; RAG: retrieval-augmented generation. (b) Mean five-benchmark accuracy and complete training time at step $100$ over Qwen3-4B and Qwen3-8B on one H100 80GB. Training configurations and timing details: Appendix~\ref{app:cost}.}
  \label{fig:intro-concept}
  \label{fig:feasibility}
\end{figure}

These informative signals can also be obtained efficiently through indexed corpus queries. For example, Infini-gram reports approximately $20$\,ms per $n$-gram count query on the $1.4$-trillion-token RedPajama corpus~\citep{liu2024infini}. This efficiency supports dense sentence-level rewards without repeated neural verification. Because corpus counts are noisy, we combine these rewards with response-level correctness supervision: corpus rewards guide individual sentences, while the correctness reward keeps training directed toward accurate final answers.

We therefore propose \textbf{CorVer}, a lightweight approach to process supervision during RL post-training for closed-book factual QA. A $0.5$B extractor identifies subject--object pairs, and an Infini-gram corpus index~\citep{liu2024infini} supplies their co-occurrence counts. CorVer incorporates these corpus rewards into RL alongside response-level correctness and format rewards. The extractor and corpus index are used only to compute rewards during training. At inference, the model answers using only the question, without corpus access. Figure~\ref{fig:feasibility}(b) shows that CorVer combines higher accuracy with substantially shorter training time on Qwen3-4B and Qwen3-8B~\citep{yang_qwen3_2025}.

Our main contributions are:
\begin{itemize}
  \item We introduce a corpus-based process reward that maps subject--object co-occurrence counts to sentence-level rewards for factual-QA post-training, without requiring a neural verifier.
  \item We demonstrate consistent accuracy gains across six models and five factual-QA benchmarks, with further gains after training on two multi-hop QA datasets. FActScore evaluation also shows more factually supported reasoning on Llama-3.2-3B-Instruct.
  \item We show that CorVer reduces complete training time by $5.5$--$10.4\times$ relative to FSPO~\citep{NEURIPS2025_ddd50f29}, RLFH~\citep{wen2025policy}, FoRAG~\citep{cai2024forag}, and KnowRL~\citep{ren_knowrl_2026} on Qwen3-4B and Qwen3-8B while achieving higher mean accuracy.
\end{itemize}
\section{Method}
\label{sec:method}

\begin{figure*}[t]
  \centering
  \includegraphics[width=\textwidth]{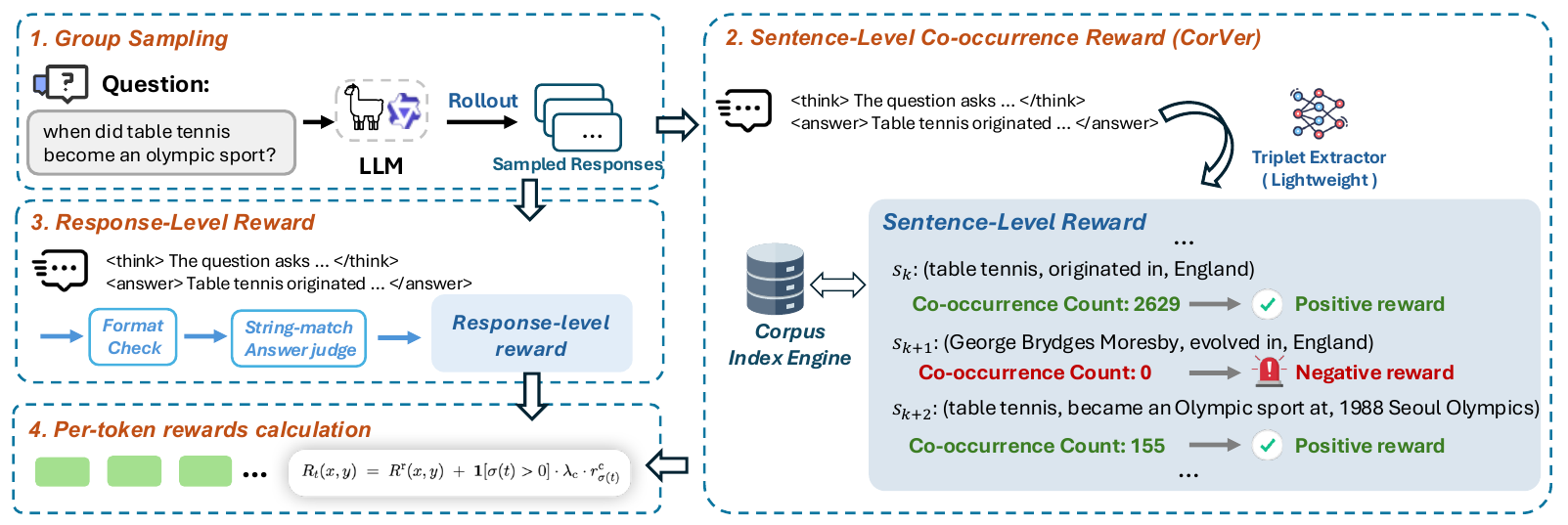}
  \caption{\textbf{CorVer pipeline.} Each sentence is scored for corpus co-occurrence via an Infini-gram index. The per-sentence reward is mapped to token-level returns through $\sigma$ and combined with response-level signals in a policy-gradient update using group relative policy optimization (GRPO).}
  \label{fig:method-overview}
\end{figure*}

CorVer trains a factual-QA policy by combining sentence-level corpus rewards with response-level correctness and format rewards. Figure~\ref{fig:method-overview} illustrates reward construction and policy optimization.

\subsection{Response Representation}
\label{sec:method-formulation}

Let $x$ denote a factual question and $\pi_\theta$ the policy. The policy receives a fixed instruction and the question. Reference answers and aliases enter only the reward computation. We sample $G$ completions per question, with $G=16$ in our experiments. A completion $y=(y_1,\ldots,y_T)$ is prompted to enclose its reasoning in \texttt{<think>...</think>} and its final answer in \texttt{<answer>...</answer>} tags. After stripping tags, we split the reasoning and final answer separately into sentences $s_1,\ldots,s_m$. The alignment $\sigma:\{1,\ldots,T\}\to\{0,1,\ldots,m\}$ maps each token to its sentence, with $\sigma(t)=0$ for tags and inter-sentence whitespace. This alignment assigns sentence-level rewards to the corresponding tokens, while response-level rewards apply throughout the completion. Training details appear in \S\ref{sec:exp-config}.

\subsection{Sentence-Level Co-occurrence Reward}
\label{sec:method-cooc}
CorVer scores each sentence based on how often its subject and object co-occur in a reference corpus. We first extract $(\text{head}, \text{relation}, \text{tail})$ triplets using a frozen $0.5$B extractor. We retain the first valid triplet and use its head and tail as the entity pair, discarding the relation. A valid pair has nonempty, non-pronominal entities whose content words form a query with at least two distinct terms. Appendix~\ref{app:aggregation} describes this \textsc{First} rule, its failure handling, and comparisons with \textsc{Min} and \textsc{RelCheck}. For sentence $s_i$, let $e_i^{\mathrm{s}}$ and $e_i^{\mathrm{o}}$ denote the retained head and tail strings. We reduce each entity to its content words to accommodate surface-form variation across the corpus. The union of these words forms a word-level AND query to the co-occurrence index, giving the count
\begin{equation}
  c_i \;=\; \mathrm{count}_{\mathcal{W}}\!\left(w_1 \,\wedge\, \cdots \,\wedge\, w_k\right),
\end{equation}
where $\mathcal{W}$ is the reference corpus (a fixed Wikipedia snapshot in our experiments) and $w_1,\ldots,w_k$ are the distinct content words derived from $(e_i^{\mathrm{s}}, e_i^{\mathrm{o}})$. The Infini-gram index~\citep{liu2024infini} counts corpus positions where the query terms co-occur within a fixed token-distance bound. Appendix~\ref{app:implementation} provides the query and extractor settings. We map this count to a sentence-level reward:
\Needspace{10\baselineskip}
\begin{equation}
\label{eq:cooc-reward}
  r_i^{\mathrm{c}} \;=\;
  \begin{cases}
    \alpha_0, & c_i = 0, \\
    \alpha_1, & 0 < c_i < \tau_1, \\
    \alpha_2, & \tau_1 \le c_i < \tau_2, \\
    \alpha_3, & c_i \ge \tau_2, \\
    0, & c_i = \bot,
  \end{cases}
\end{equation}
where $\alpha_0 < \alpha_1 \le \alpha_2 < \alpha_3$ are bounded reward levels and $\tau_1 < \tau_2$ are integer thresholds. We set $c_i=\bot$ when no valid entity pair or usable count is available, distinguishing missing evidence from a successful query with zero matches. The observation in Figure~\ref{fig:calibration-curve} motivates using counts as a weak factuality signal (Appendix~\ref{app:human-eval}). The bounded mapping supplies sentence rewards alongside correctness supervision. Section~\ref{sec:exp-config} gives reward values and thresholds.

\subsection{Response-Level Rewards}
\label{sec:method-channels}

CorVer combines the sentence-level signal with two response-level rewards. The \emph{correctness reward} $R^{\mathrm{j}}(x,y)$ evaluates only the final answer outside the reasoning block through deterministic string and alias matching. Incorrect, refused, and empty answers receive the same penalty. The \emph{format reward} $R^{\mathrm{f}}(y)$ checks for a nonempty, closed final-answer block and includes a linear penalty for excess completion length. Reward values and format rules appear in Appendix~\ref{app:reward-config}.

\subsection{Corpus Rewards and Token Advantages}
\label{sec:method-stepwise}

For completion $y^{(g)}$, let $r_{g,i}^{\mathrm c}$ denote the reward of sentence $i$. The alignment $\sigma_g$ assigns this reward to the corresponding tokens. With $r_{g,0}^{\mathrm c}=0$, the token return is
\begin{equation}
\label{eq:token-reward}
\begin{aligned}
  R_{g,t}
  &= \underbrace{\lambda_{\mathrm j}R^{\mathrm j}(x,y^{(g)})
     +\lambda_{\mathrm f}R^{\mathrm f}(y^{(g)})}_{\text{response-level supervision}} \\
  &\quad +\underbrace{\lambda_{\mathrm c}\mathbf1[\sigma_g(t)>0]\,
     r^{\mathrm c}_{g,\sigma_g(t)}}_{\text{sentence-level corpus supervision}}.
\end{aligned}
\end{equation}
The weights $\lambda_{\mathrm j},\lambda_{\mathrm f},\lambda_{\mathrm c}$ control the three reward channels. The first two terms form the response-level return $R^{\mathrm r}$ shown in Figure~\ref{fig:method-overview}. Tokens in either the reasoning or answer block receive the reward of their sentence, while tags and inter-sentence whitespace receive only response-level supervision.

We normalize token returns using group statistics computed from each completion's mean return:
\begin{equation}
\label{eq:token-advantage}
  \bar R_g=\frac{1}{T_g}\sum_{t=1}^{T_g}R_{g,t},
  \qquad
  A_{g,t}=\frac{R_{g,t}-\mu_x}{s_x+\varepsilon},
\end{equation}
where $T_g$ counts valid completion tokens, excluding padding, and $\mu_x$ and $s_x$ are the mean and standard deviation of $\{\bar R_g\}_{g=1}^{G}$ for question $x$, with $\varepsilon=10^{-4}$. The numerator retains sentence-level reward differences, so sentences within a completion can receive different advantages. Setting $\lambda_{\mathrm c}=0$ yields a shared response-level advantage within each completion. We use these advantages in clipped GRPO with Kullback--Leibler (KL) regularization against the frozen initial base model~\citep{shao_deepseekmath_2024}. Low-rank adaptation (LoRA)~\citep{hu_lora_2021} updates average over valid tokens within each completion and then across the group (Appendix~\ref{app:algorithm}).

\paragraph{Efficient implementation.}
Sentences from multiple completions are processed together, repeated queries reuse cached counts, and the resulting rewards are assigned back to each completion's tokens. Extraction runs in a persistent worker with CUDA Graph execution, and token-reward tensors are constructed on the central processing unit (CPU) and transferred to the GPU in batches. Appendix~\ref{app:implementation} lists the corpus query parameters and sampling settings for the extractor.
\raggedbottom
\section{Experimental Setup}
\label{sec:exp-config}

\textbf{Data and model.} We use TriviaQA~\citep{joshi2017triviaqa}, NQ-Open~\citep{kwiatkowski2019natural}, PopQA~\citep{mallen2023not}, SimpleQA~\citep{wei2024measuring}, and TruthfulQA~\citep{lin2022truthfulqa}. The main comparison uses Llama-3.2-3B-Instruct, Llama-3.1-8B-Instruct~\citep{grattafiori_llama_2024}, Qwen3-4B, and Qwen3-8B~\citep{yang_qwen3_2025}, with additional tests on OLMo-2-13B-Instruct~\citep{olmo_2_2025} and Qwen3-14B. We also use HotpotQA~\citep{yang2018hotpotqa} and 2WikiMultiHopQA~\citep{ho2020twowikimultihopqa} for multi-hop QA. We evaluate Llama-3.2-3B-Instruct reasoning on $1{,}000$ HotpotQA validation questions.

\textbf{Baseline and training.} As discussed above, existing methods typically compute sentence-level factuality rewards through NLI verification or LLM-as-judge. FSPO~\citep{NEURIPS2025_ddd50f29}, our closest comparator, combines NLI-based sentence rewards with answer correctness. Related methods target complementary goals: RLFH~\citep{wen2025policy} balances truthfulness and informativeness, FoRAG~\citep{cai2024forag} targets factual retrieval-augmented long-form QA, and KnowRL~\citep{ren_knowrl_2026} encourages knowledge-boundary awareness and abstention. Their original goals differ from closed-book answer accuracy, so we evaluate their transfer to this setting. Raw is the unmodified model. Training uses single-GPU LoRA and one NQ-Open/WebQuestions~\citep{berant2013semantic} pool shared across methods and all six base models. CorVer uses GRPO. In Eq.~\eqref{eq:cooc-reward}, we set $(\alpha_0,\alpha_1,\alpha_2,\alpha_3)=(-0.2,-0.1,0,+0.1)$ and $(\tau_1,\tau_2)=(5,20)$. Multi-hop training uses each target dataset separately. We test author-reported hyperparameters (FSPO, RLFH), rollout allocation (FoRAG), and training duration (KnowRL), retaining the stronger complete baseline row by macro accuracy. Appendices~\ref{app:baselines}, \ref{app:multihop}, and~\ref{app:baseline-comparisons} provide training and configuration details.

\textbf{Evaluation.} We report closed-book answer accuracy and its five-benchmark macro average. For multi-hop QA, exact match (EM) requires a normalized reference match, while token-level F1, the harmonic mean of token precision and recall, credits partial matches. FActScore-NPM~\citep{min2023factscore} uses a nonparametric masked language model (NPM) to evaluate reasoning through micro atomic precision, supported facts per response, coverage, and facts per response. Training cost is complete run time on one H100 80GB. Protocols appear in Appendices~\ref{app:datasets}, \ref{app:multihop}, and~\ref{app:cost}.

\begin{table}[!t]
  \caption{Factual QA accuracy (\%) on five benchmarks. For FSPO, RLFH, FoRAG, and KnowRL, we retain the complete row with higher macro accuracy where alternative settings were tested. CorVer uses its final step-$100$ checkpoint. Bold/underline mark the best/second-best scores within each model. Differences are relative to the best competitor (\textcolor{gain}{red}: higher; \textcolor{loss}{green}: lower).}
  \label{tab:main}
  \centering
  \small
  \resizebox{\textwidth}{!}{%
  \begin{tabular}{llcccccc}
    \toprule
    Model & Method & TriviaQA & NQ-Open & PopQA & SimpleQA & TruthfulQA & Macro\ \\
    \midrule
    \multirow{6}{*}{Llama-3.2-3B-Instruct}
      & Raw & 57.84 & 36.65 & 18.54 & 2.24 & 36.35 & 30.32 \\
      & FoRAG & \underline{60.79} & \underline{40.28} & 21.17 & 2.24 & \underline{42.35} & 33.37 \\
      & RLFH & 59.12 & 36.90 & 20.75 & 2.61 & 37.21 & 31.32 \\
      & FSPO & 59.45 & 40.25 & \underline{22.11} & \textbf{4.30} & 41.13 & \underline{33.45} \\
      & KnowRL & 50.23 & 29.70 & 15.17 & 1.53 & 27.42 & 24.81 \\
      & \textbf{CorVer (ours)} & \textbf{66.19}\dg{+5.40} & \textbf{45.48}\dg{+5.20} & \textbf{26.26}\dg{+4.15} & \underline{3.91}\dl{-0.39} & \textbf{46.76}\dg{+4.41} & \textbf{37.72}\dg{+4.27} \\
    \midrule
    \multirow{6}{*}{Llama-3.1-8B-Instruct}
      & Raw & 74.67 & 44.93 & 31.97 & 5.69 & 46.27 & 40.70 \\
      & FoRAG & 66.79 & 33.43 & 25.73 & 5.55 & 36.23 & 33.55 \\
      & RLFH & 72.49 & 41.22 & 29.92 & 5.36 & 38.92 & 37.58 \\
      & FSPO & \underline{79.81} & \underline{51.02} & \textbf{36.77} & \textbf{6.89} & \underline{47.61} & \underline{44.42} \\
      & KnowRL & 71.14 & 41.52 & 29.19 & 2.43 & 36.60 & 36.18 \\
      & \textbf{CorVer (ours)} & \textbf{81.50}\dg{+1.69} & \textbf{53.07}\dg{+2.05} & \underline{36.69}\dl{-0.08} & \underline{6.70}\dl{-0.19} & \textbf{49.45}\dg{+1.84} & \textbf{45.48}\dg{+1.06} \\
    \midrule
    \multirow{6}{*}{Qwen3-4B}
      & Raw & 52.32 & 24.82 & 17.73 & 2.96 & \underline{45.29} & 28.62 \\
      & FoRAG & 51.52 & 24.71 & 16.93 & 2.10 & 39.53 & 26.96 \\
      & RLFH & 50.68 & 24.60 & 15.99 & 1.60 & 36.60 & 25.89 \\
      & FSPO & \underline{53.55} & \underline{26.79} & \underline{18.14} & \underline{3.37} & 44.80 & \underline{29.33} \\
      & KnowRL & 44.09 & 20.69 & 11.35 & 0.46 & 27.42 & 20.80 \\
      & \textbf{CorVer (ours)} & \textbf{57.57}\dg{+4.02} & \textbf{31.86}\dg{+5.07} & \textbf{22.16}\dg{+4.02} & \textbf{4.23}\dg{+0.86} & \textbf{50.31}\dg{+5.02} & \textbf{33.22}\dg{+3.89} \\
    \midrule
    \multirow{6}{*}{Qwen3-8B}
      & Raw & 63.75 & 30.94 & 21.45 & 3.01 & 42.72 & 32.37 \\
      & FoRAG & 60.86 & 28.42 & 19.81 & 2.73 & 41.00 & 30.57 \\
      & RLFH & 62.77 & 30.39 & 20.52 & 2.91 & 42.35 & 31.79 \\
      & FSPO & \underline{64.77} & \underline{32.24} & \underline{22.77} & \underline{3.91} & \underline{49.33} & \underline{34.60} \\
      & KnowRL & 62.19 & 30.78 & 18.20 & 2.45 & 38.80 & 30.48 \\
      & \textbf{CorVer (ours)} & \textbf{67.39}\dg{+2.62} & \textbf{36.87}\dg{+4.63} & \textbf{27.57}\dg{+4.80} & \textbf{4.92}\dg{+1.01} & \textbf{51.65}\dg{+2.32} & \textbf{37.68}\dg{+3.08} \\
    \bottomrule
  \end{tabular}
  }
\end{table}

\section{Experiments}
\label{sec:exp}

\subsection{Main Results}
\label{sec:exp-main}

CorVer leads in $17$ of $20$ model--benchmark settings and exceeds FSPO, the strongest baseline by macro accuracy, on all four models (Table~\ref{tab:main}). Both Qwen3 models outperform FSPO on every benchmark, whereas the Llama models show a less uniform advantage. Llama-3.1-8B has the smallest macro margin over FSPO at $1.06$ percentage points (pp). These results suggest limited transfer from the related baselines' original objectives to closed-book accuracy. KnowRL illustrates the mismatch: longer Qwen3-4B training increases refusals, which receive no accuracy credit, without improving accuracy (Appendix~\ref{app:baseline-comparisons}).

Qualitative analysis of Llama-3.1-8B finds newly correct answers alongside regressions, sometimes involving familiar background knowledge that diverts responses from the requested fact. These losses partly offset the corrections, whereas newly correct answers substantially outnumber regressions in other models. We conjecture that this model's existing knowledge and reasoning behavior are less well matched to corpus rewards, contributing to its smaller advantage over FSPO (\S\ref{sec:exp-scaling}).

\subsection{Gains across Model Families and Scales}
\label{sec:exp-scaling}

\begin{figure}[t]
  \centering
  \includegraphics[width=0.98\textwidth]{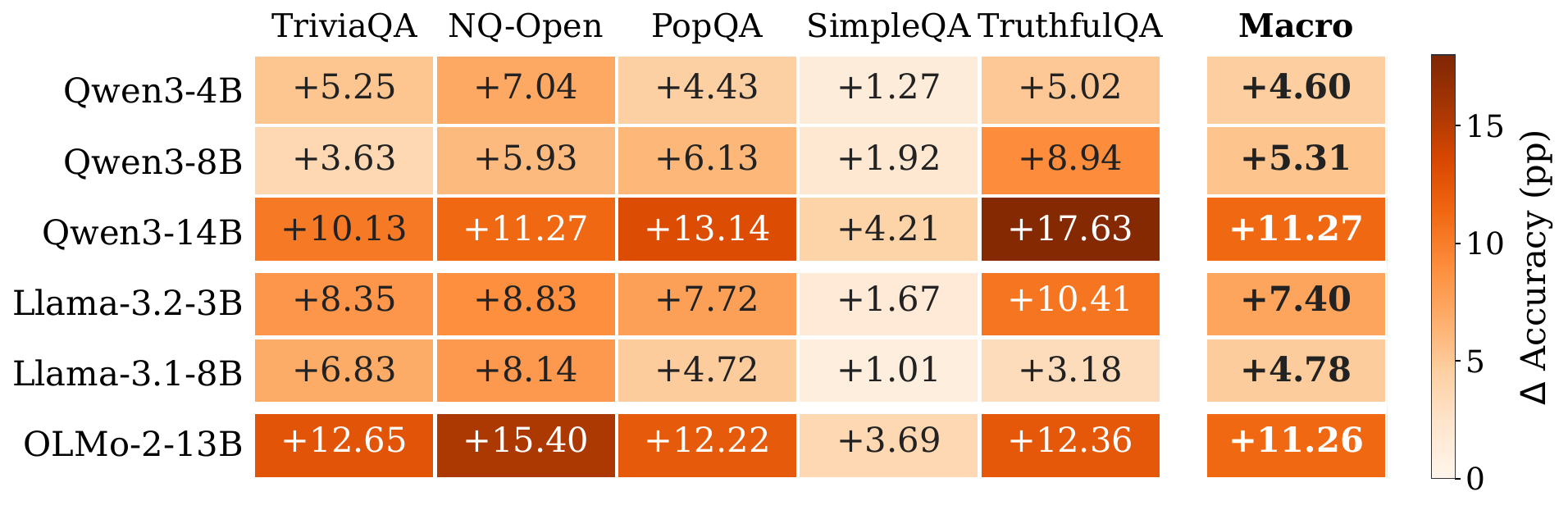}
  \caption{\textbf{CorVer accuracy gains over Raw (pp).} Rows group models by family. Macro is the mean gain across the five benchmarks.}
  \label{fig:scaling}
\end{figure}

Figure~\ref{fig:scaling} covers six models from three families, with positive gains in all $30$ model--benchmark settings. Mean gains are largest on TruthfulQA ($+9.59$ pp) and NQ-Open ($+9.44$ pp), followed by PopQA ($+8.06$ pp) and TriviaQA ($+7.81$ pp), and smallest on SimpleQA ($+2.29$ pp). The task-level pattern also differs between models of similar size: at $8$B, Llama gains more on NQ-Open, whereas Qwen3 gains more on TruthfulQA. This reversal suggests that the benefit depends on the combination of model and task, which the macro average alone does not reveal.

The largest gains occur on OLMo-2-13B and Qwen3-14B: both exceed $11$ pp, substantially more than any $3$B--$8$B model. Within Qwen3, gains rise from $+4.60$ to $+5.31$ and $+11.27$ pp as scale increases. This pattern is consistent with our hypothesis that a stronger pretrained foundation makes imperfect corpus rewards more useful (\S\ref{sec:calibration}). We conjecture that richer knowledge helps a model select and connect facts it already knows, whereas smaller models may be more susceptible to misleading rewards. The Llama trend differs ($+7.40$ pp at $3$B versus $+4.78$ pp at $8$B), suggesting that this benefit depends on existing capabilities as well as scale. Appendix~\ref{app:full-results} reports the full accuracies.

\subsection{Accuracy and Training Cost}
\label{sec:exp-cost}

{\setlength{\intextsep}{4pt}
\begin{table}[H]
  \setlength{\belowcaptionskip}{6pt}
  \caption{Accuracy and complete training cost at step $100$, averaged over Qwen3-4B and Qwen3-8B. Time ratios are relative to CorVer.}
  \label{tab:cost-summary}
  \centering
  \small
  \setlength{\tabcolsep}{10pt}
  \begin{tabular}{lrrr}
    \toprule
    Method & Macro accuracy (\%) $\uparrow$ & Time (h:mm) $\downarrow$ & Time ratio $\downarrow$ \\
    \midrule
    \textbf{CorVer} & \textbf{35.45} & \textbf{1:49} & \textbf{1.00}$\times$ \\
    FSPO & 31.97 & 10:01 & 5.52$\times$ \\
    RLFH & 28.84 & 18:53 & 10.41$\times$ \\
    FoRAG & 28.76 & 14:05 & 7.76$\times$ \\
    KnowRL & 25.64 & 15:55 & 8.77$\times$ \\
    \bottomrule
  \end{tabular}
\end{table}
}

We report complete training wall-clock time, excluding job queueing and evaluation. This includes application programming interface (API) communication and response waiting, since policy updates require the remote judges' rewards. CorVer gains $3.49$ pp over FSPO while taking less than one fifth as long (Table~\ref{tab:cost-summary}). CorVer is also faster than FSPO and FoRAG, whose rewards are computed locally without remote API calls (Appendix~\ref{app:cost}).

\Needspace{16\baselineskip}
\subsection{Factuality of the Generated Reasoning}
\label{sec:exp-factscore}

\begin{wraptable}{r}{0.49\textwidth}
  \vspace{-0.75\baselineskip}
  \centering
  \caption{FActScore-NPM for Llama-3.2-3B-Instruct. Precision is micro-averaged.}
  \label{tab:factscore}
  \small
  \setlength{\tabcolsep}{5pt}
  \resizebox{\linewidth}{!}{%
  \begin{tabular}{lrr}
    \toprule
    Metric & Raw & CorVer \\
    \midrule
    Atomic precision (\%) $\uparrow$ & 83.73 & \textbf{84.98} \\
    Supported facts / response $\uparrow$ & 1.292 & \textbf{1.443} \\
    Coverage (\%) $\uparrow$ & 61.40 & \textbf{67.90} \\
    \midrule
    Facts / response & 1.543 & 1.698 \\
    \bottomrule
  \end{tabular}
  }
\end{wraptable}

Higher answer accuracy does not establish that the reasoning itself is more factual. We therefore compare Raw and CorVer using FActScore-NPM, which assesses support for atomic facts: individual, checkable claims extracted from the reasoning. Atomic precision is the fraction of extracted facts supported by the evaluator's evidence, pooled across responses. Supported facts per response measures the amount of supported content, while coverage is the fraction of responses with at least one scorable fact. Facts per response counts all extracted facts, including unsupported ones.

CorVer produces more facts per response while increasing atomic precision (Table~\ref{tab:factscore}), indicating that the additional content is accompanied by better factual support. Higher coverage further shows that a larger share of responses contains scorable facts.

\subsection{Learning on Multi-Hop QA}
\label{sec:exp-multihop}

Multi-hop QA tests whether sentence-level corpus rewards can improve answers that require combining multiple facts. We compare task-specific CorVer and FSPO training against Raw on HotpotQA and 2WikiMultiHopQA.

\begin{table}[H]
  \caption{Closed-book multi-hop QA after dataset-specific training, evaluated by answer EM and F1 (\%). Bold marks the best score within each model.}
  \label{tab:multihop}
  \centering
  \small
  \setlength{\tabcolsep}{6pt}
  \begin{tabular*}{\textwidth}{@{\extracolsep{\fill}}llrrrr@{}}
    \toprule
    & & \multicolumn{2}{c}{HotpotQA} & \multicolumn{2}{c}{2WikiMultiHopQA} \\
    \cmidrule(lr){3-4}\cmidrule(lr){5-6}
    Model & Method & EM $\uparrow$ & F1 $\uparrow$ & EM $\uparrow$ & F1 $\uparrow$ \\
    \midrule
    \multirow{3}{*}{Llama-3.2-3B-Instruct}
      & Raw & 7.99 & 14.28 & 3.03 & 7.28 \\
      & FSPO & 14.80 & 22.00 & 18.69 & 22.20 \\
      & \textbf{CorVer} & \textbf{16.42} & \textbf{23.06} & \textbf{19.61} & \textbf{23.47} \\
    \midrule
    \multirow{3}{*}{Qwen3-4B}
      & Raw & 10.93 & 18.16 & 12.53 & 20.76 \\
      & FSPO & 17.19 & \textbf{24.78} & 16.86 & 19.88 \\
      & \textbf{CorVer} & \textbf{17.85} & 24.71 & \textbf{25.25} & \textbf{29.27} \\
    \midrule
    \multirow{3}{*}{Llama-3.1-8B-Instruct}
      & Raw & 14.44 & 23.69 & 8.87 & 17.71 \\
      & FSPO & 18.87 & 27.48 & 24.33 & 30.01 \\
      & \textbf{CorVer} & \textbf{22.77} & \textbf{30.56} & \textbf{24.82} & \textbf{30.13} \\
    \midrule
    \multirow{3}{*}{Qwen3-8B}
      & Raw & 19.01 & 26.51 & 21.62 & 26.19 \\
      & FSPO & 21.11 & 29.24 & 19.48 & 22.72 \\
      & \textbf{CorVer} & \textbf{21.42} & \textbf{29.91} & \textbf{27.50} & \textbf{32.01} \\

    \bottomrule
  \end{tabular*}
\end{table}

CorVer achieves higher EM than FSPO in every model--dataset comparison (Table~\ref{tab:multihop}), indicating that it produces the complete reference answer more often. Its advantage is especially pronounced on 2WikiMultiHopQA, where both Qwen models gain over $8$ pp in EM over FSPO. The F1 results broadly agree: CorVer leads except on Qwen3-4B HotpotQA, where the methods differ by only $0.07$ pp. Together, these results show that sentence-level co-occurrence rewards can support training for questions requiring multiple facts, even though the reward does not explicitly verify the links between reasoning steps. Appendix~\ref{app:multihop} reports additional Raw--CorVer comparisons on the larger models.
\section{Analysis and Discussion}
\label{sec:analysis}

\subsection{Contributions of the Reward Components}
\label{sec:exp-ablation}

The ablations separate the reward source from its assignment. We hold the base model fixed at Llama-3.2-3B-Instruct and remove one component at a time. A1 removes corpus rewards, while A2--A4 retain the zero-count penalty of $-0.2$. Across all five benchmarks, the full configuration outperforms each variant and Raw (Table~\ref{tab:ablation}).

\begin{table}[H]
  \setlength{\belowcaptionskip}{6pt}
  \caption{Ablation accuracy (\%) on Llama-3.2-3B-Instruct. Each variant removes the named component. Bold marks the best score on each benchmark.}
  \label{tab:ablation}
  \centering
  \small
  \setlength{\tabcolsep}{4pt}
  \resizebox{\textwidth}{!}{%
  \begin{tabular}{lccccc}
    \toprule
    Variant                                & TriviaQA & NQ-Open & PopQA & SimpleQA & TruthfulQA \\
    \midrule
    Raw & 57.84 & 36.65 & 18.54 & 2.24 & 36.35 \\
    \textbf{CorVer (full)} & \textbf{66.19} & \textbf{45.48} & \textbf{26.26} & \textbf{3.91} & \textbf{46.76} \\
    \midrule
    \quad A1: $-$ corpus reward (vanilla GRPO; correctness $+$ format rewards) & 59.61 & 39.81 & 20.10 & 2.73 & 40.51 \\
    \quad A2: $-$ correctness reward (per-token corpus reward $+$ format only) & 59.02 & 40.06 & 21.11 & 2.52 & 39.78 \\
    \quad A3: $-$ per-token alignment (sentence-mean corpus reward) & 63.53 & 43.88 & 24.67 & 3.56 & 41.74 \\
    \quad A4: $-$ self-filter (full $9{,}680$-question pool) & 63.52 & 42.96 & 23.09 & 2.64 & 43.33 \\
    \bottomrule
  \end{tabular}
  }
\end{table}

Corpus rewards and the correctness reward contribute differently across benchmarks. Removing the corpus reward (A1) lowers accuracy on all five, including TriviaQA ($66.19$ to $59.61$) and NQ-Open ($45.48$ to $39.81$). Removing the correctness reward (A2) produces the largest drop on TriviaQA ($7.17$ pp), SimpleQA ($1.39$ pp), and TruthfulQA ($6.98$ pp), whereas removing the corpus reward costs more on NQ-Open ($5.67$ pp) and PopQA ($6.16$ pp). Reward assignment also matters. A3 retains the same corpus signal but averages it into a single response score. Accuracy falls on every benchmark, most on TruthfulQA ($41.74$ versus $46.76$), suggesting that the model benefits from rewards tied to individual sentences. Finally, using the unfiltered question pool (A4) lowers accuracy throughout, with the largest drops on TruthfulQA ($3.43$ pp) and PopQA ($3.17$ pp).

\subsection{Gains across Entity Popularity}
\label{sec:exp-popularity}

\begin{figure}[H]
  \centering
  \includegraphics[width=\textwidth]{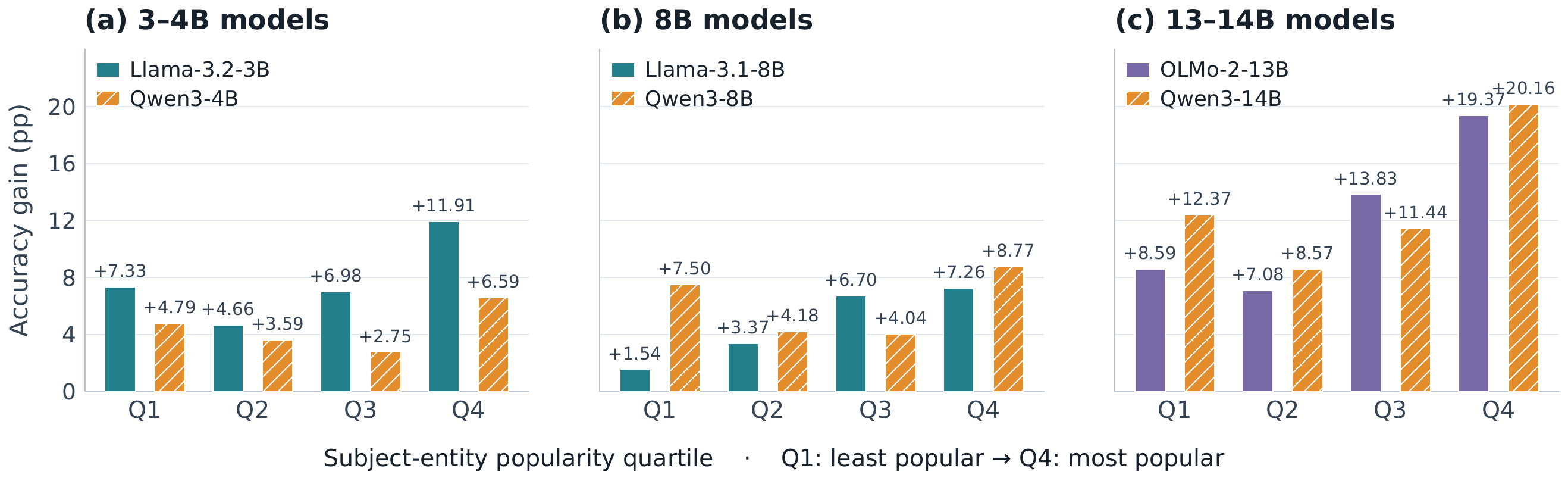}
  \caption{PopQA accuracy gains over Raw (pp) by entity popularity. All models use the same quartiles, from the least popular entities (Q1) to the most popular (Q4).}
  \label{fig:popularity}
\end{figure}

We examine how CorVer's PopQA gains vary with subject-entity popularity~\citep{mallen2023not}. Figure~\ref{fig:popularity} covers six models in three scale groups ($3$--$4$B, $8$B, and $13$--$14$B), retaining each model's results within the group. All models use shared popularity quartiles over the full $14{,}267$ questions; Appendix~\ref{app:popularity} reports each model's full-set gain and the gains within all four quartiles.

CorVer achieves its largest gains in Q4 for every model. The equally weighted six-model means are $+12.34$ pp for Q4, $+7.02$ and $+7.62$ pp for Q1 and Q3, and $+5.24$ pp for Q2.

These averages show that CorVer's gains are strongest for popular entities but also substantial for the least popular entities, rather than increasing monotonically with popularity. The ordering varies across models: Qwen3-4B and Qwen3-8B gain less in Q3 than in Q2. We conjecture that low-popularity entities benefit from penalties on unsupported associations, while high-popularity entities benefit from reinforcement of correct associations. At intermediate popularity, co-occurrence rewards may distinguish correct from incorrect associations less clearly, potentially limiting the mean gain in Q2.

\subsection{Why Corpus Rewards Help in Post-Training}
\label{sec:calibration}

Corpus counts can penalize rare facts or reward false claims involving familiar entities. Their usefulness in CorVer rests on three considerations. First, our $700$-sentence audit is stratified by co-occurrence count (Appendix~\ref{app:human-eval}). Within these sampled buckets, $81\%$ of sentences with $c_i \geq 20$ are correct, while zero-count sentences are $76\%$ incorrect and $24\%$ correct. These bucket-level proportions support positive and negative rewards as noisy signals, with factual precision increasing monotonically across all five buckets (Figure~\ref{fig:calibration-curve}).

Second, response-level correctness rewards complement corpus rewards when frequency and correctness disagree. Correct answers receive a correctness reward even when rare associations incur corpus penalties, while incorrect answers receive a correctness penalty despite frequent entity pairs. Removing either reward source reduces accuracy across all five benchmarks (\S\ref{sec:exp-ablation}), supporting their complementary roles.

Third, we hypothesize that LLM-as-judge supervision may involve a form of circular reasoning. Prior work links factual recall to pretraining exposure and documents difficulties with rarely co-occurring entities~\citep{kandpal2023large,kang2023impact}. If the policy and judge learn from similar corpora, limited exposure to the same facts may impair both generation and verification. When the judge relies on its parametric knowledge, the potential circularity lies in using one model's uncertain knowledge to validate another's, without independent factual evidence. Under these conditions, LLM-as-judge rewards may not be more reliable than noisy corpus rewards. CorVer queries corpus statistics directly, without requiring a verifier to recall the relevant facts. Although these counts also miscredit some sentences, they provide an external signal to refine a policy with existing factual and reasoning capabilities. KL regularization discourages large departures from the initial policy, which may help preserve useful behavior under noisy rewards. We conjecture that this combination helps explain the observed training gains.
\section{Related Work}
\label{sec:related}

\paragraph{Process supervision and corpus evidence.}
RL post-training commonly uses response-level objectives~\citep{ouyang_training_2022,shao_deepseekmath_2024,deepseek-ai_deepseek-r1_2025}, while process supervision evaluates intermediate reasoning~\citep{lightman_lets_2023,nie2026stop}. Fine-Grained RLHF~\citep{wu2023finegrained} learns segment-level rewards for reinforcement learning from human feedback (RLHF). Retrieved evidence supports generation in RAG~\citep{NEURIPS2020_6b493230} and factuality evaluation in FActScore~\citep{min2023factscore}, while Self-RAG~\citep{asai2024selfrag} combines retrieval with self-critique. Factual recall correlates with co-occurrence and pretraining exposure~\citep{kang2023impact,kandpal2023large}, and QuCo~\citep{min2025quco} uses co-occurrence to guide retrieval. CorVer instead uses Infini-gram counts~\citep{liu2024infini} as inexpensive process rewards.

\paragraph{Factuality-aware post-training.}
FSPO~\citep{NEURIPS2025_ddd50f29} combines sentence-level entailment checks with answer rewards, making it our closest baseline. RLFH~\citep{wen2025policy} balances truthfulness and informativeness: its policy judges atomic facts against evidence, with information rewards discouraging uninformative refusals or overly brief answers. FoRAG~\citep{cai2024forag} targets factual, coherent long-form answers grounded in retrieved evidence through outline-based generation and fine-grained reward modeling. KnowRL~\citep{ren_knowrl_2026} aims to keep reasoning within the model's knowledge boundaries, rewarding refusal when uncertain alongside correctness and factuality. Other work studies factuality-oriented online RL~\citep{chen2025learning}, direct preference optimization~\citep{tian2024factuality}, factuality-aware supervised and preference training~\citep{lin2024flame}, and refusal-aware instruction tuning~\citep{zhang2024rtuning}. CorVer focuses on efficient sentence-level rewards for closed-book QA, combining corpus counts with answer correctness.
\section{Limitations and Conclusion}
\label{sec:conclusion}

\phantomsection
\label{sec:limitations}
\paragraph{Limitations.}
Corpus co-occurrence is an imperfect proxy for factuality: frequent entity pairs can occur in false claims, and rare but correct facts can receive penalties. Its usefulness depends on corpus coverage and extraction quality, while systematic mismatches between frequency and correctness may limit further gains. Our evaluation is strictly closed-book: policies receive only the question at inference. CorVer's training-time corpus queries return counts for rewards, without supplying retrieved text to the policy. We do not compare against RAG systems that condition answer generation on retrieved passages.

\paragraph{Conclusion.}
CorVer turns inexpensive corpus co-occurrence statistics into sentence-level process rewards for factual-QA post-training. Across six models, it improves accuracy in all $30$ model--benchmark settings, with additional gains in task-specific multi-hop QA and atomic factual precision. On Qwen3-4B and Qwen3-8B, it achieves higher macro accuracy than all four RL baselines while reducing complete training time by $5.5$--$10.4\times$. Ablations support both combining corpus and correctness rewards and preserving rewards at the sentence level. The results establish corpus statistics as a practical source of process rewards for factual QA, reducing the cost of sentence-level supervision while improving both final answers and the factual content of reasoning. The extractor and index are needed only during training, so these gains require no additional verification components at inference.

\subsection*{AI use statement}

LLM-based tools were used for code and implementation assistance and for manuscript writing and refinement, including editing and LaTeX formatting. The authors reviewed the AI-assisted code and manuscript and take responsibility for the final research design, implementation, analyses, reported results, and paper content.

\subsection*{Ethics statement}

The manual factuality audit was conducted by the authors; no external annotators, crowdworkers, or personal data were involved.

CorVer aims to improve factuality, but generated answers can still be incorrect. Corpus co-occurrence may reflect biases in source coverage and underrepresent rare entities and facts. Models trained with these rewards should therefore not be treated as authoritative fact-checkers, and their outputs require independent verification in high-stakes applications.

\subsection*{Reproducibility statement}

The public repository provides the CorVer training code, model configurations, the selected training questions, and links to publicly available baseline resources: \url{https://github.com/shichengf/CorVer}. Appendix~\ref{app:setup} details the training, data processing, and evaluation protocols.


\flushbottom
\bibliography{custom}
\bibliographystyle{iclr2027/iclr2027_conference}

\appendix

\section{Training Setup and Recipe}
\label{app:setup}

\subsection{Comparison Protocol and Training Settings}
\label{app:baselines}

All methods are evaluated in closed-book QA: the policy receives only the question at inference. Evidence retrieval and reward models are used during training where required by each method.

\subsubsection{Methods and Their Implementations}

\paragraph{CorVer.}
CorVer combines corpus-based sentence rewards with correctness and format rewards. A frozen $0.5$B extractor maps sentences to entity pairs, and an Infini-gram index supplies their co-occurrence counts. The resulting rewards are assigned to sentence tokens for GRPO training. The extractor runs on the training GPU, while corpus lookup runs on the CPU.

\paragraph{FSPO.}
FSPO~\citep{NEURIPS2025_ddd50f29}\footnote{\url{https://github.com/nusnlp/FSPO}} combines answer correctness with factuality rewards on reasoning steps. A resident Hughes Hallucination Evaluation Model (HHEM-2.1) NLI verifier checks sentences against evidence, and its factuality rewards enter token-level advantages for GRPO training.

\paragraph{RLFH.}
RLFH~\citep{wen2025policy}\footnote{\url{https://github.com/AlignRM/RLFH}} uses truthfulness and informativeness rewards for proximal policy optimization (PPO) training. Our frozen judge is \texttt{Qwen/Qwen3.8-Flash-Next-FP8} ($180$B total parameters, $6$B active), accessed through a managed OpenAI-compatible endpoint. Its prompt and decoding configuration follow the original RLFH setup.

\paragraph{FoRAG.}
FoRAG~\citep{cai2024forag} decomposes generated content into subclaims and verifies them against evidence to compute fine-grained factuality rewards. We implement this reward procedure for closed-book policy training with PPO. Both FoRAG and RLFH maintain a critic and reuse each sampled batch for four PPO epochs.

\paragraph{KnowRL.}
KnowRL~\citep{ren_knowrl_2026}\footnote{\url{https://github.com/zjunlp/KnowRL}} combines answer correctness with atomic-fact support to encourage awareness of knowledge boundaries. Our GRPO implementation uses GPT-4o-mini, DeBERTa-v3 NLI, and a Wikipedia knowledge base for reward computation. We retain its reward for appropriate abstention (Appendix~\ref{app:knowrl-training}).

\newpage
\subsubsection{Shared Experimental Conditions}

The main five-benchmark comparison holds the following conditions fixed across methods for each base model:
\begin{itemize}
  \item \textbf{Training questions.} All methods and all six base models use the same selected pool of $6{,}621$ questions from NQ-Open and WebQuestions, prepared once. Appendix~\ref{app:datasets} gives the selection procedure.
  \item \textbf{Adaptation and hardware.} Training uses single-GPU LoRA with rank $128$, scaling $256$, and seed $42$, with one seed per configuration. Baseline adapters use dropout $0$ and no bias. Each timed run uses one H100 80GB.
  \item \textbf{Generation limit.} Training rollouts and main-table evaluation responses are limited to $1{,}000$ generated tokens.
  \item \textbf{Evaluation.} Methods share the same question-only system prompt, benchmark questions, answer extraction, scoring rules, and decoding settings for a given base model. Empty answers and refusals receive zero accuracy credit. The prompt, model-specific decoding settings, and scoring rules are given in Appendix~\ref{app:datasets}.
\end{itemize}

\begin{table}[H]
  \caption{Shared rollout allocations and additional baseline configurations, grouped by model. Within each main model, the $100$-step rows with $C=16$ share the same generation budget. FoRAG (adjusted) and KnowRL (extended) report additional configurations. FSPO and RLFH use their main-table settings. $Q$ is questions per step and $C$ is completions per question. Steps are optimizer updates for GRPO and outer batches for PPO. Responses count newly generated completions ($Q\times C\times\text{steps}$).}
  \label{tab:rollout-config}
  \centering
  \footnotesize
  \begin{tabular*}{\linewidth}{@{\extracolsep{\fill}}llrrrr@{}}
    \toprule
    Base model & Method / configuration & $Q$ & $C$ & Steps & Responses \\
    \midrule
    \multirow[t]{7}{*}{Llama-3.2-3B} & CorVer & 3 & 16 & 100 & $4{,}800$ \\
     & FSPO & 3 & 16 & 100 & $4{,}800$ \\
     & RLFH & 3 & 16 & 100 & $4{,}800$ \\
     & FoRAG (initial) & 3 & 16 & 100 & $4{,}800$ \\
     & FoRAG (adjusted) & 4 & 1 & 100 & $400$ \\
     & KnowRL & 3 & 16 & 100 & $4{,}800$ \\
     & KnowRL (extended) & 3 & 16 & 200 & $9{,}600$ \\
    \midrule
    \multirow[t]{7}{*}{Qwen3-4B} & CorVer & 4 & 16 & 100 & $6{,}400$ \\
     & FSPO & 4 & 16 & 100 & $6{,}400$ \\
     & RLFH & 4 & 16 & 100 & $6{,}400$ \\
     & FoRAG (initial) & 4 & 16 & 100 & $6{,}400$ \\
     & FoRAG (adjusted) & 4 & 1 & 100 & $400$ \\
     & KnowRL & 4 & 16 & 100 & $6{,}400$ \\
     & KnowRL (extended) & 4 & 16 & 200 & $12{,}800$ \\
    \midrule
    \multirow[t]{7}{*}{Llama-3.1-8B} & CorVer & 3 & 16 & 100 & $4{,}800$ \\
     & FSPO & 3 & 16 & 100 & $4{,}800$ \\
     & RLFH & 3 & 16 & 100 & $4{,}800$ \\
     & FoRAG (initial) & 3 & 16 & 100 & $4{,}800$ \\
     & FoRAG (adjusted) & 4 & 1 & 100 & $400$ \\
     & KnowRL & 3 & 16 & 100 & $4{,}800$ \\
     & KnowRL (extended) & 3 & 16 & 200 & $9{,}600$ \\
    \midrule
    \multirow[t]{7}{*}{Qwen3-8B} & CorVer & 3 & 16 & 100 & $4{,}800$ \\
     & FSPO & 3 & 16 & 100 & $4{,}800$ \\
     & RLFH & 3 & 16 & 100 & $4{,}800$ \\
     & FoRAG (initial) & 3 & 16 & 100 & $4{,}800$ \\
     & FoRAG (adjusted) & 4 & 1 & 100 & $400$ \\
     & KnowRL & 3 & 16 & 100 & $4{,}800$ \\
     & KnowRL (extended) & 3 & 16 & 200 & $9{,}600$ \\
    \midrule
    OLMo-2-13B & CorVer & 3 & 16 & 100 & $4{,}800$ \\
    \midrule
    Qwen3-14B & CorVer & 3 & 16 & 100 & $4{,}800$ \\
    \bottomrule
  \end{tabular*}
\end{table}

\subsubsection{Method-Specific Settings and Selection}

\begin{table}[H]
  \caption{Algorithm, learning rate, KL coefficient, and reward components. CorVer uses KL $0.01$ for Qwen3-4B, $0.003$ for Llama-3.1-8B, and $0.001$ otherwise. RLFH uses KL $0.01$ on Llama-3.2-3B and Qwen3-4B, and $0.05$ on the two $8$B models. Other baseline learning-rate (LR) and KL values apply across the four main models. Configuration comparisons and checkpoint selection are described in Appendix~\ref{app:baseline-comparisons}.}
  \label{tab:baseline-config}
  \centering
  \footnotesize
  \setlength{\tabcolsep}{3pt}
  \begin{tabularx}{\linewidth}{@{}lllll>{\raggedright\arraybackslash}X@{}}
    \toprule
    Method & Algo.\ & $G$ & Steps / ckpt.\ & LR / $\beta_{\mathrm{KL}}$ & Reward-side components \\
    \midrule
    CorVer & GRPO & $16$ & $100$ / step-$100$ & $1 \times 10^{-5}$ / $0.001$ & $0.5$B extractor $+$ Infini-gram lookup \\
    FSPO   & GRPO & $16$ & $100$ / step-$100$ & $2 \times 10^{-6}$ / $0.04$ & HHEM-2.1 NLI, threshold $0.5$ \\
    RLFH   & PPO  & --- & $100$ / step-$100$ & $5 \times 10^{-7}$ / $0.01$ or $0.05$ & Qwen3.8-Flash-Next-FP8 judge \\
    FoRAG  & PPO  & --- & $100$ / step-$100$ & $5 \times 10^{-7}$ / $0.02$ & subclaim decomposition $+$ evidence-field verification~\citep{cai2024forag} \\
    KnowRL & GRPO & $16$ & $100$ / step-$100$ & $1 \times 10^{-5}$ / $0.001$ & GPT-4o-mini judge $+$ DeBERTa-v3 NLI (thr.\ $0.3$) $+$ Wikipedia knowledge base \\
    \bottomrule
  \end{tabularx}
\end{table}

Learning rates and KL coefficients are tuned empirically using the original papers and available official implementations as references. Tables~\ref{tab:rollout-config} and~\ref{tab:baseline-config} give the generation budgets and selected settings. Where alternatives are evaluated, we retain the complete baseline row with higher five-benchmark macro accuracy, with all scores from one checkpoint. PPO epochs reuse responses and do not multiply the generation budget. Complete-run costs are reported in Appendix~\ref{app:cost}.

\paragraph{Selected baseline configurations.}
FSPO uses the adapted configuration at rollout temperature $0.7$. RLFH uses author-key settings at $1.0$ for Llama-3.2-3B and Qwen3-4B, and adapted settings at $0.7$ for the two $8$B models. FoRAG uses adjusted rollout allocations for Llama-3.2-3B and Qwen3-8B and initial allocations for the other two models, all at temperature $1.0$. KnowRL uses step-$100$ checkpoints for accuracy and cost comparisons. Appendix~\ref{app:baseline-comparisons} reports the configuration and duration comparisons supporting these choices.

\paragraph{CorVer training and rewards.}
\label{app:reward-config}
CorVer trains a fresh LoRA adapter from the official base model without a supervised fine-tuning (SFT) cold start, using rollout temperature $1.0$ and the settings in Tables~\ref{tab:rollout-config} and~\ref{tab:baseline-config}. For Eq.~\eqref{eq:cooc-reward}, $(\alpha_0,\alpha_1,\alpha_2,\alpha_3)=(-0.2,-0.1,0.0,+0.1)$ and $(\tau_1,\tau_2)=(5,20)$ (Appendix~\ref{app:calibration}). Correctness gives $+2$ for a matched final answer and $-1$ for an incorrect, refused, or empty answer. We set $\lambda_{\mathrm f}=\lambda_{\mathrm c}=1$ and $\lambda_{\mathrm j}=1$, except $\lambda_{\mathrm j}=2$ for Llama-3.1-8B. Under token averaging, the mean corpus reward lies in $[-0.2,0.1]$, below the correctness reward swing of $3\lambda_{\mathrm j}$. Alternative aggregation rules appear in Appendix~\ref{app:aggregation}.

The format reward checks the final-answer block outside the reasoning block and penalizes excess completion length:
\begin{equation}
\label{eq:format-length-reward}
R^{\mathrm f}(y)=
\begin{cases}
+1, & \text{the final-answer block is nonempty and closed},\\
-1, & \text{otherwise}
\end{cases}
-\eta\frac{\max\{0,L(y)-L_0\}}{L_0},
\end{equation}
where $L(y)$ is the completion length in tokens, $\eta=0.5$, and $L_0=512$. Answer correctness is scored separately, so a well-formed incorrect answer can receive the positive format component.

\subsection{Datasets and Evaluation Protocol}
\label{app:datasets}

\paragraph{Evaluation datasets.}
Each CorVer run evaluates the final checkpoint on five closed-book QA datasets in series: TriviaQA validation ($17{,}944$ questions), NQ-Open validation ($3{,}610$), PopQA test ($14{,}267$), SimpleQA ($4{,}326$), and TruthfulQA ($817$). Each TriviaQA record carries a question, a primary answer, and an alias list.

\paragraph{Training data and selection.}
Training questions come from NQ-Open and WebQuestions. Preparation removes duplicate questions within the training pool and applies Wikipedia entity grounding. The $6{,}621$-question pool is selected once using Llama-3.1-8B and shared across all methods and all six base models, without separate selection for each model or method. The budgets in Table~\ref{tab:rollout-config} count sampled instances, which need not be distinct or cover the pool.

The pool combines learning-zone questions, which received both correct and incorrect sampled answers during selection, with already-mastered anchor questions. Ablation A4 uses the unfiltered candidate pool to examine this selection (\S\ref{sec:exp-ablation}).

\paragraph{Training and evaluation grading.}
For CorVer's five-benchmark experiments, the training correctness reward uses normalized answer and alias matching on the final answer outside the reasoning block. At evaluation, TriviaQA, NQ-Open, PopQA, and SimpleQA use bidirectional substring matching after normalization: a response matches if its final answer contains a reference answer or alias, or is contained in one. TruthfulQA instead computes ROUGE-1 F1 between the final answer and each reference, taking separate maxima over the correct and incorrect reference sets. An answer is correct only when the maximum for correct references is strictly greater than the maximum for incorrect references. Ties, including two zero maxima, score $0$. We report the accuracy of this binary decision, not mean ROUGE-1 F1. Appendix~\ref{app:scoring-checks} reports supplementary checks using one-way matching and semantic judgments.

On all five benchmarks, empty answers and explicit refusals score $0$, and the denominator retains every test question, including duplicates inherent in the source data. A response counts as an explicit refusal only when its normalized answer span exactly matches a refusal phrase (\emph{I don't know}, \emph{I do not know}, \emph{I have no comment}). Empty answers are distinguished from explicit refusals.

\paragraph{Reasoning-trace factuality evaluation.}
\label{app:factscore-metrics}
We compare Llama-3.2-3B-Instruct Raw with CorVer ($-0.2$, \textsc{First}, final step-$100$) on the same $1{,}000$ HotpotQA distractor-validation questions, sampled in advance with seed $42$. Both receive only the same system prompt and question. Generation uses the Brain Floating Point 16-bit (BF16) format, greedy decoding, a $1{,}000$-token output limit, and a $2{,}048$-token context limit. We score the \texttt{<think>} content preceding the first \texttt{<answer>}; if think tags are absent, we use the text before the answer block. Final answers are excluded.

We use FActScore-NPM~\citep{min2023factscore} with independent Wikipedia retrieval and a QA-oriented article-retrieval adaptation. A Qwen3 extractor produces verbatim atomic claims at temperature $0$ with thinking disabled; exact duplicates are removed. Article retrieval uses the fixed \texttt{enwiki-20230401.db} Wikipedia snapshot. SQLite Full-Text Search 5 (FTS5) retrieves the top article using an OR query over up to $12$ deduplicated non-stopwords from the claim, with Best Matching 25 (BM25) title/body weights $5/1$. Official BM25 passage retrieval selects the top three passages within that article. The official \texttt{facebook/npm-single} model scores support in 32-bit floating-point (FP32) precision; a fact is supported only if its NPM probability is strictly greater than $0.3$. The evaluator uses no HotpotQA context, supporting facts, or gold answers, and applies no $\gamma$ length penalty. Its retrieval and support scoring are separate from the training co-occurrence lookup.

\paragraph{Factuality metrics.}
For each model, let $N=1{,}000$ be the number of responses, $F$ the total atomic facts, $S$ the supported facts, and $A$ the responses with at least one scorable fact. We report micro atomic precision $100S/F$, supported facts per response $S/N$, coverage $100A/N$, and facts per response $F/N$. Empty reasoning and fact-free responses remain in $N$. Coverage measures scorable-response prevalence, and facts per response is a quantity diagnostic. Raw has $(F,S,A)=(1{,}543,1{,}292,614)$ and CorVer has $(1{,}698,1{,}443,679)$.

\paragraph{Prompting and generation length.}
All main-table methods use the following system prompt, which requests brief reasoning, a direct answer, the specified abstention format, and no repetition. The user message contains only the original question, without retrieved material or examples. The shared generation limit is $1{,}000$ tokens.
\begin{lstlisting}[language={},showstringspaces=false,keepspaces=true]
Answer using this exact format:
<think>brief reasoning</think>
<answer>direct answer</answer>

If you genuinely do not know, reply exactly: <think>unknown</think><answer>I don't know</answer>
Do not loop or repeat the same point or phrase.
\end{lstlisting}
\paragraph{Decoding settings.}
\label{app:qwen-decoding}
All methods share evaluation decoding within each base model. Llama-3.2-3B-Instruct and Qwen3-8B use greedy decoding with temperature $0$ and repetition penalty $1.0$. Llama-3.1-8B-Instruct uses the same settings with repetition penalty $1.02$. Its Raw and both FSPO configurations use one beginning-of-sequence (BOS) token per input. Qwen3-4B follows the model card's recommended thinking-mode sampling: temperature $0.6$, top-$p$ $0.95$, top-$k$ $20$, presence penalty $1.5$, and repetition penalty $1.0$.\footnote{\url{https://huggingface.co/Qwen/Qwen3-4B\#best-practices}} In the scaling experiments, Qwen3-14B and OLMo-2-13B-Instruct use greedy decoding with temperature $0$ and repetition penalty $1.0$ for Raw and CorVer. Other baselines are not evaluated on these two models. The shared $1{,}000$-token output limit applies throughout.

\subsection{Multi-Hop Dataset-Specific Training and Evaluation}
\label{app:multihop}

\paragraph{Training data and initialization.}
For the CorVer and FSPO runs in Tables~\ref{tab:multihop} and~\ref{tab:multihop-larger}, each model--dataset run initializes a fresh LoRA adapter from the official base model. All six models use the same target-task pools: $4{,}064$ HotpotQA and $1{,}208$ 2WikiMultiHopQA questions. The pools were selected using a response-level-reward model and are fixed across all six models. We check their official training-split provenance, remove duplicate training questions, and exclude question-identifier or normalized-question overlaps with either official development set. Reference answers enter only the training reward; reference passages, titles, supporting facts, and evidence are not supplied to the policy.

\paragraph{CorVer training configuration.}
CorVer retains the Llama-3.2-3B co-occurrence, format, and length rewards (Appendix~\ref{app:reward-config}). Correctness uses official normalized exact match ($+2$ if correct, $-1$ otherwise), including official aliases and demonyms for 2WikiMultiHopQA. Runs use LoRA rank/scaling $128/256$, dropout $0$, learning rate $10^{-5}$, KL coefficient $0.001$, AdamW8bit, BF16, $16$ generations per question, and seed $42$. The additional $4$B--$14$B runs use H100 GPUs. Evaluation uses the final checkpoint after $100$ optimizer steps.

\paragraph{FSPO comparison.}
FSPO is trained separately on both target datasets for the four $3$B--$8$B models, yielding eight runs with the same selected pools as CorVer. All use LoRA rank/scaling $128/256$ and $3$ questions with $16$ responses each per step, including Qwen3-4B, totaling $4{,}800$ responses over $100$ steps. Final step-$100$ checkpoints are evaluated on each full development set. These four models form the method comparison, while the larger models provide supplementary Raw--CorVer evaluations.

\paragraph{Evaluation protocol.}
We evaluate the full official development sets: $7{,}405$ HotpotQA and $12{,}576$ 2WikiMultiHopQA questions. Within each model, Raw, CorVer, and FSPO (where evaluated) share the closed-book system prompt and question, with fixed prompt date 07 Sep 2026. Decoding and the $1{,}000$-token output limit follow Appendix~\ref{app:qwen-decoding}. Llama-3.2-3B uses context length $2{,}048$. Official scoring scripts compute answer EM and token-overlap F1, with refused and empty answers scoring zero and all development questions retained in the denominator. Tables~\ref{tab:multihop} and~\ref{tab:multihop-larger} report answer scores, while reasoning FActScore is evaluated separately in \S\ref{sec:exp-factscore}.

\paragraph{Additional results on larger models.}
Table~\ref{tab:multihop-larger} additionally reports Raw and CorVer results on OLMo-2-13B-Instruct and Qwen3-14B after the same dataset-specific training.

\begin{table}[htbp]
  \caption{Additional closed-book multi-hop results on larger models (\%). CorVer uses the final step-$100$ checkpoint trained separately for each dataset. Answer EM and F1 cover the full official development sets; CorVer uses a zero-count penalty of $-0.2$. Bold marks the higher value within each model pair.}
  \label{tab:multihop-larger}
  \centering
  \small
  \setlength{\tabcolsep}{6pt}
  \begin{tabular*}{\textwidth}{@{\extracolsep{\fill}}llrrrr@{}}
    \toprule
    & & \multicolumn{2}{c}{HotpotQA} & \multicolumn{2}{c}{2WikiMultiHopQA} \\
    \cmidrule(lr){3-4}\cmidrule(lr){5-6}
    Model & Method & EM $\uparrow$ & F1 $\uparrow$ & EM $\uparrow$ & F1 $\uparrow$ \\
    \midrule
    \multirow{2}{*}{OLMo-2-13B-Instruct}
      & Raw & 20.23 & 27.73 & 21.04 & 24.86 \\
      & \textbf{CorVer} & \textbf{25.73} & \textbf{35.28} & \textbf{26.49} & \textbf{32.17} \\
    \midrule
    \multirow{2}{*}{Qwen3-14B}
      & Raw & 16.91 & 22.49 & 11.73 & 14.34 \\
      & \textbf{CorVer} & \textbf{23.07} & \textbf{31.50} & \textbf{26.18} & \textbf{31.26} \\
    \bottomrule
  \end{tabular*}
\end{table}

\subsection{Implementation Details}
\label{app:implementation}
\raggedbottom
\afterpage{\flushbottom}

\paragraph{Fixed index and extractor.}
Following~\citet{min2025quco}, we implement the word-level AND query as a conjunctive normal form (CNF) count over corpus token positions using Infini-gram~\citep{liu2024infini}. The fixed Wikipedia index uses \texttt{max\_diff\_tokens=1000} and \texttt{max\_clause\_freq=500000}. The distance limit is measured in tokens of the fixed Llama-2 index tokenizer; it is an engine distance bound, not a symmetric $\pm1{,}000$-token window or a document count. The training co-occurrence index is separate from the knowledge base used for NPM evaluation.

We use QuCo-extractor-0.5B~\citep{min2025quco}, a Qwen2.5-0.5B-Instruct model fine-tuned for triplet extraction, and keep it frozen during policy training. The extractor worker uses BF16, temperature $0.7$, top-$p$ $0.8$, top-$k$ $20$, and repetition penalty $1.1$, with at most $256$ new tokens and a total context limit of $4{,}096$.

\paragraph{Efficient reward computation.}
\label{app:efficient-reward}
The policy uses vLLM for rollout generation with eager execution. We batch sentences across sampled completions and extract triplets with a persistent, frozen QuCo-extractor-0.5B worker using CUDA Graph execution on the same GPU. Extracted triplets are mapped back to their original sentences before corpus queries and reward assignment.

A local corpus index and a query-count cache reduce repeated lookups. Successful counts, including zero, are cached, while failed queries are not. We construct token rewards on the CPU using the sentence--token alignment and transfer them to the GPU in batches, excluding padding from the objective. Each sentence reward applies only to its corresponding tokens. Ablation A3 instead broadcasts an averaged sentence reward (\S\ref{sec:exp-ablation}).

At inference, only the trained policy is used, with no extractor or corpus lookup.

\section{Motivating Audit of Corpus Co-occurrence}
\label{app:human-eval}
\label{app:calibration}

We manually examined $700$ sentences sampled by co-occurrence bucket from Llama-3.1-8B-Instruct responses to questions drawn from publicly available QA sources (Figure~\ref{fig:calibration-curve}). Five of the authors independently labeled the $400$ sentences in the zero-count and highest-count buckets after checking claims with web search. Four of the authors labeled the remaining $300$ sentences, with external sources available for verification. Final labels were determined by majority vote, with $2$--$2$ ties resolved through discussion. Labels assess sentence factuality independently of final-answer correctness.

\section{Co-occurrence Reward Design and Analysis}
\label{app:reward-design}

Appendices~\ref{app:penalty-sweep} and~\ref{app:aggregation} evaluate reward sensitivity and triplet aggregation on TriviaQA using models trained on the NQ-Open and WebQuestions pool (Appendix~\ref{app:datasets}). The main reward settings, including the zero-count penalty $-0.2$, the \textsc{First} rule, and the count thresholds, were selected independently of these evaluations. TriviaQA questions and responses are used exclusively for evaluation and analysis.

\subsection{Zero-Count Penalty Sensitivity}
\label{app:penalty-sweep}

\begin{figure}[H]
  \centering
  \begin{minipage}[t]{0.49\textwidth}
    \vspace{0pt}
    \centering
    \includegraphics[width=\linewidth]{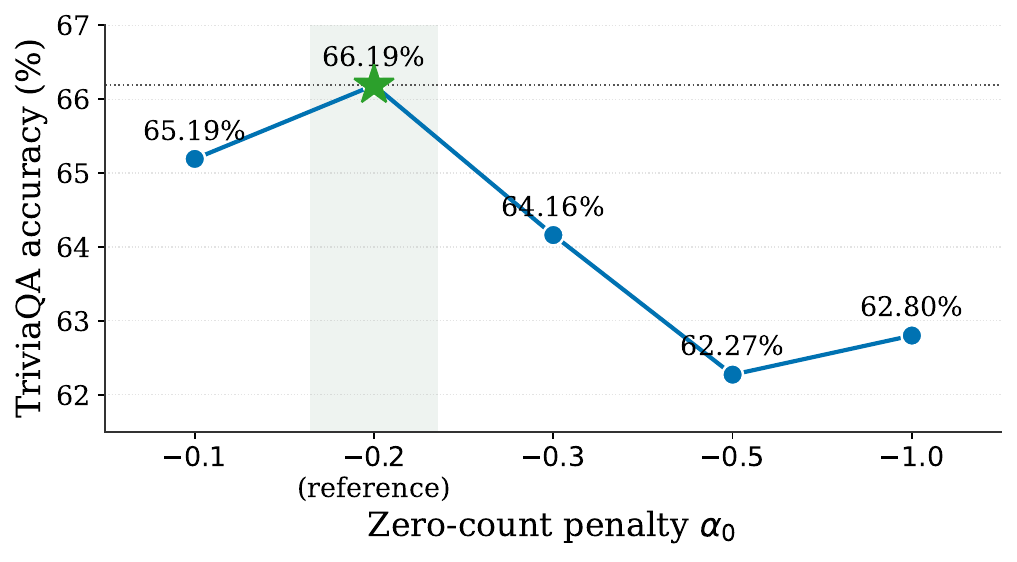}
    \caption{TriviaQA accuracy under five zero-count penalties. The star marks the $-0.2$ setting, and the dotted line shows its reference accuracy.}
    \label{fig:penalty-sweep}
  \end{minipage}\hfill
  \begin{minipage}[t]{0.47\textwidth}
    \vspace{0pt}
    We vary only the zero-count penalty $\alpha_0$ over $\{-0.1,-0.2,-0.3,-0.5,-1.0\}$ using Llama-3.2-3B-Instruct, the \textsc{First} scoring rule, seed $42$, and final step-$100$ checkpoints. Evaluation covers all $17{,}944$ TriviaQA validation questions.

    \medskip
    The $-0.2$ setting reaches $66.19\%$, the highest accuracy among the five settings. Relative to this reference, $-0.1$, $-0.3$, $-0.5$, and $-1.0$ reduce accuracy by $1.00$, $2.03$, $3.92$, and $3.39$ pp, respectively. Differences are computed from correct-answer counts.
  \end{minipage}
\end{figure}

\subsection{Triplet Aggregation and Relation Checks}
\label{app:aggregation}

This comparison uses Llama-3.2-3B-Instruct, seed $42$, final step-$100$ checkpoints, and zero-count penalty $-0.2$. Full uses the \textsc{First} rule and matches the main experiment.

\paragraph{Valid triplets and queries.}
A valid triplet is a three-element list whose subject and object are nonempty strings outside the implementation's pronoun list and can form an AND query with at least two distinct content terms.

\paragraph{\textsc{First} and \textsc{Min}.}
\textsc{First} queries the first valid triplet's subject and object. An unsuccessful query gives reward $0$, without trying a later triplet. \textsc{Min} queries all valid pairs and uses the smallest successful count. Failed queries are omitted, while a successful zero count remains eligible for the minimum. If no usable count is available, the reward is $0$.

\paragraph{\textsc{RelCheck}.}
\textsc{RelCheck} starts from \textsc{First}. A subject--object count $c_i\ge20$ triggers a subject--relation--object query. A successful zero count lowers the reward from $+0.1$ to $-0.05$. A positive count, a failed query, or an unusable relation query leaves the original reward unchanged. The same content-word extraction is applied to all three fields, and the merged query must contain at least three distinct terms. This check adds relation words to the corpus query rather than applying an entailment model.

\paragraph{Query construction.}
Content-word extraction first selects capitalized non-stopwords longer than one character, then falls back to non-stopwords longer than two characters, and finally to the original stripped string. Query terms retain their original order and are deduplicated case-insensitively. The shared extractor and these query rules are fixed across the three variants.

\paragraph{Comparison results.}
On the full TriviaQA validation set, Full (\textsc{First}) achieves $66.19\%$ accuracy, compared with $61.25\%$ for \textsc{Min} and $61.34\%$ for \textsc{RelCheck}, corresponding to drops of $4.94$ and $4.85$ pp. In this comparison, checking more triplets or adding relation terms does not improve answer accuracy. For \textsc{Min}, sparse coverage of secondary claims may introduce extra penalties.

\begin{table}[H]
  \caption{Triplet-aggregation comparison on Llama-3.2-3B-Instruct (TriviaQA validation, $N=17{,}944$). All runs use seed $42$, final step-$100$ checkpoints, and zero-count penalty $-0.2$. Differences are computed from correct-answer counts.}
  \label{tab:aggregation}
  \centering
  \small
  \setlength{\tabcolsep}{10pt}
  \begin{tabular}{lrr}
    \toprule
    Configuration & Accuracy (\%) & $\Delta$ vs. Full (pp) \\
    \midrule
    \textbf{Full (\textsc{First})} & \textbf{66.19} & $0.00$ \\
    \textsc{Min} & 61.25 & $-4.94$ \\
    \textsc{RelCheck} & 61.34 & $-4.85$ \\
    \bottomrule
  \end{tabular}
\end{table}

\subsection{CorVer Algorithm}
\label{app:algorithm}

Algorithm~\ref{alg:stepfact} summarizes how CorVer constructs token rewards and advantages. Reference answers are used only for reward computation. The group statistics are computed from the mean token return of each completion, with padding excluded throughout.

\begin{algorithm}[h]
\caption{CorVer: Corpus Rewards with Token Alignment}
\label{alg:stepfact}
\begin{algorithmic}[1]
\Require Policy $\pi_\theta$, training data $\mathcal D$, group size $G$
\For{each training iteration}
  \State Sample question $x$ and reference answers from $\mathcal D$
  \State Sample $\{y^{(g)}\}_{g=1}^{G}\sim\pi_\theta(\cdot\mid x)$
  \For{$g=1,\ldots,G$}
    \State Split reasoning and final answer separately into sentences $s_{g,1:m_g}$
    \For{$i=1,\ldots,m_g$}
      \State Extract the first valid entity pair from $s_{g,i}$ and query its count $c_{g,i}$
      \State Use $c_{g,i}=\bot$ if no valid pair or usable count is available
      \State $r_{g,i}^{\mathrm c}\gets$ Eq.~\eqref{eq:cooc-reward}
    \EndFor
    \State Compute $R^{\mathrm j}(x,y^{(g)})$ and $R^{\mathrm f}(y^{(g)})$
    \State Build alignment $\sigma_g$ and token returns $R_{g,t}$ via Eq.~\eqref{eq:token-reward}
    \State $\bar R_g\gets T_g^{-1}\sum_{t=1}^{T_g}R_{g,t}$ over valid tokens
  \EndFor
  \State $\mu_x\gets\operatorname{mean}_g(\bar R_g)$, $s_x\gets\operatorname{std}_g(\bar R_g)$
  \State $A_{g,t}\gets(R_{g,t}-\mu_x)/(s_x+10^{-4})$
  \State Update LoRA with the clipped GRPO objective and KL to the frozen initial base model
\EndFor
\end{algorithmic}
\end{algorithm}

\newpage
\raggedbottom
\subsection{Qualitative Zero-Count Cases}
\label{app:case-study}

Figure~\ref{fig:case-study} shows three single-sentence errors penalized by zero corpus counts but accepted by GPT-4o-mini. Below, two complete Llama-3.1-8B-Instruct CorVer responses to NQ-Open questions have correct final answers despite flawed reasoning. Sentence IDs link each response to the extracted triplets, counts, rewards, and factual assessments.

\definecolor{errcol}{HTML}{A01B1B}
\definecolor{mutedcol}{HTML}{606060}

\begin{figure}[H]
  \centering
  \small
  \setlength{\tabcolsep}{5pt}
  \renewcommand{\arraystretch}{1.15}
  \begin{tabularx}{\linewidth}{@{}>{\raggedright\arraybackslash}p{0.24\linewidth}X@{}}
    \toprule
    Error type & Generated claim and indexed entity pair \\
    \midrule
    \textbf{A. Digit substitution} & Ronkonkoma's ZIP code is \textcolor{errcol}{\textbf{11777}}. \quad Gold: 11779.\newline
    Indexed pair: (Ronkonkoma, 11777). \\
    \addlinespace[5pt]
    \textbf{B. Entity substitution} & \textcolor{errcol}{\textbf{Jennifer Coolidge}} appears in \emph{Grown Ups}. \quad Gold: Maria Bello.\newline
    Indexed pair: (Jennifer Coolidge, Grown Ups). \\
    \addlinespace[5pt]
    \textbf{C. Near-name substitution} & Ella Fitzgerald's mother was \textcolor{errcol}{\textbf{Height Fitzgerald}}.\newline
    Indexed pair: (Ella Fitzgerald, Height). \\
    \bottomrule
  \end{tabularx}
  \par\smallskip
  \textbf{All three cases:} count $=0$ \quad CorVer: reject ($r_i^{\mathrm c}=-0.2$) \quad GPT-4o-mini: accept
  \caption{Three single-sentence errors. Zero corpus co-occurrence penalizes each fabricated entity association, while GPT-4o-mini accepts the claim.}
  \label{fig:case-study}
\end{figure}

\paragraph{Complete responses and sentence rewards.}
Case~1 contains a fabricated name and an incorrect detail outside the first triplet. Case~2 contains a fabricated category association. Both tables separate factual assessment from co-occurrence rewards.

\begin{casetrace}{Case 1. Philadelphia Flyers}
\textbf{Question:} when was the last time philly won the stanley cup\par
\textbf{Reference answer:} 1975

\medskip
\textbf{Generated reasoning}\par
\textsuperscript{[1]}The Philadelphia Flyers, a professional ice hockey team based in Philadelphia, Pennsylvania, won the Stanley Cup in 1975. \textsuperscript{[2]}The team defeated the Buffalo Sabres in the Stanley Cup Finals, winning the series 4--2 with a game 5 victory on May 19, 1975. \textsuperscript{[3]}\textcolor{errcol}{\textbf{Philipp Mehldau}} was the Governor of Pennsylvania during this time.

\smallskip
\textbf{Generated final answer}\par
\textsuperscript{[4]}The Philadelphia Flyers last won the Stanley Cup in 1975. \textsuperscript{[5]}The Stanley Cup was last won by the Philadelphia Flyers in 1975.

\medskip
\textbf{Sentence rewards}\par\smallskip
\setlength{\tabcolsep}{4pt}
\renewcommand{\arraystretch}{1.12}
\begin{tabularx}{\linewidth}{@{}l>{\raggedright\arraybackslash}X>{\raggedleft\arraybackslash}p{0.12\linewidth}>{\raggedleft\arraybackslash}p{0.07\linewidth}>{\raggedright\arraybackslash}p{0.18\linewidth}@{}}
\toprule
ID & Extracted triplet & Count $c_i$ & $r_i^{\mathrm c}$ & Assessment \\
\midrule
1 & (Philadelphia Flyers, won, Stanley Cup) & $12{,}323$ & $+0.1$ & Correct \\
\addlinespace[4pt]
2 & (the team, defeated, Buffalo Sabres) & $16{,}484$ & $+0.1$ & Correct triplet\textsuperscript{*} \\
\addlinespace[4pt]
3 & (Philipp Mehldau, Governor of, Pennsylvania) & $0$ & $\textcolor{errcol}{-0.2}$ & \textcolor{errcol}{Fabricated} \\
\addlinespace[4pt]
4 & (Philadelphia Flyers, last won, Stanley Cup) & $12{,}323$ & $+0.1$ & Correct \\
\addlinespace[4pt]
5 & (Stanley Cup, last won by, Philadelphia Flyers) & $12{,}323$ & $+0.1$ & Correct \\
\bottomrule
\end{tabularx}

\smallskip
\textbf{Reading the trace.} Sentences 1, 4, and 5 correctly identify the Flyers' 1975 Cup, which remains their most recent. Sentence 3 invents a governor's name. The governor was Milton Shapp.

\smallskip
\textsuperscript{*}\textbf{Coverage limitation.} The Flyers defeated the Sabres 4--2, so sentence 2's extracted triplet is correct. Its game number and date are wrong: the deciding game was Game~6 on May~27, rather than Game~5 on May~19. These details fall outside the first extracted triplet. Checking secondary triplets could expose additional errors when those details are extracted. Appendix~\ref{app:aggregation} defines and compares these alternative rules.
\end{casetrace}

\begin{casetrace}{Case 2. National Collegiate Athletic Association (NCAA) basketball championship}
\textbf{Question:} who won the ncaa basketball championship in 1989\par
\textbf{Reference answer:} Michigan

\medskip
\textbf{Generated reasoning}\par
\textsuperscript{[1]}Michigan was led by Glen Rice, who was named the Most Outstanding Player. \textsuperscript{[2]}Michigan won the championship by defeating Seton Hall 80--79. \textsuperscript{[3]}The game took place at the King Dome in Seattle, Washington. \textsuperscript{[4]}The 1989 NCAA Men's Division~I Basketball Championship was the 61st edition of the NCAA \textcolor{errcol}{\textbf{Field Hockey}} Championships and was held on April~3, 1989.

\smallskip
\textbf{Generated final answer}\par
\textsuperscript{[5]}The Michigan Wolverines won the NCAA basketball championship in 1989. \textsuperscript{[6]}The 1989 NCAA Men's Division~I Basketball Championship was held at the King Dome in Seattle, Washington.

\medskip
\textbf{Sentence rewards}\par\smallskip
\setlength{\tabcolsep}{4pt}
\renewcommand{\arraystretch}{1.12}
\begin{tabularx}{\linewidth}{@{}l>{\raggedright\arraybackslash}X>{\raggedleft\arraybackslash}p{0.12\linewidth}>{\raggedleft\arraybackslash}p{0.07\linewidth}>{\raggedright\arraybackslash}p{0.18\linewidth}@{}}
\toprule
ID & Extracted triplet & Count $c_i$ & $r_i^{\mathrm c}$ & Assessment \\
\midrule
1 & (Michigan, led by, Glen Rice) & $163$ & $+0.1$ & Correct \\
\addlinespace[4pt]
2 & (Michigan, defeated, Seton Hall) & $389$ & $+0.1$ & Correct \\
\addlinespace[4pt]
3 & (the game, took place at, King Dome) & $542$ & $+0.1$ & Correct \\
\addlinespace[4pt]
4 & (1989 NCAA Basketball Championship, edition of, NCAA Field Hockey Championships) & $0$ & $\textcolor{errcol}{-0.2}$ & \textcolor{errcol}{Fabricated} \\
\addlinespace[4pt]
5 & (Michigan Wolverines, won, NCAA basketball championship) & $3{,}500$ & $+0.1$ & Correct \\
\addlinespace[4pt]
6 & (1989 Championship, held at, King Dome) & $10$ & $0$ & Correct \\
\bottomrule
\end{tabularx}

\smallskip
\textbf{Reading the trace.} Glen Rice was the 1989 tournament's Most Outstanding Player. Michigan defeated Seton Hall 80--79 in overtime at the Seattle Kingdome. Sentence 4 fabricates a connection to field hockey and receives $-0.2$. Sentence 5 matches the reference answer. Sentence 6 is correct but receives a neutral reward because its count is $10$.
\end{casetrace}

\afterpage{\flushbottom}

\section{Results, Diagnostics, and Cost}
\label{app:results-diag}

\subsection{Full Per-Model Results}
\label{app:full-results}
\label{app:checkpoint}

Table~\ref{tab:full-results} reports five-benchmark accuracy for all six models under the shared evaluation protocol ($40{,}964$ questions per model and variant). Raw uses the official base weights, and CorVer uses the final checkpoint after $100$ optimizer steps.

\begin{table}[H]
  \caption{Full per-model evaluation accuracy ($\%$) on the five closed-book QA datasets (TriviaQA validation, NQ-Open validation, PopQA test, SimpleQA, TruthfulQA), scored under the unified answer-required protocol on the full evaluation sets. CorVer uses a zero-count penalty of $-0.2$ and step-$100$ checkpoints.}
  \label{tab:full-results}
  \centering
  \small
  \resizebox{\textwidth}{!}{%
  \begin{tabular}{llccccc}
    \toprule
    Model & Variant & TriviaQA & NQ-Open & PopQA & SimpleQA & TruthfulQA \\
    \midrule
    \multirow{2}{*}{Llama-3.2-3B-Instruct}
      & Raw & $57.84$ & $36.65$ & $18.54$ & $2.24$ & $36.35$ \\
      & CorVer & $66.19$ & $45.48$ & $26.26$ & $3.91$ & $46.76$ \\
    \cmidrule(lr){2-7}
    \multirow{2}{*}{Qwen3-4B}
      & Raw & $52.32$ & $24.82$ & $17.73$ & $2.96$ & $45.29$ \\
      & CorVer & $57.57$ & $31.86$ & $22.16$ & $4.23$ & $50.31$ \\
    \cmidrule(lr){2-7}
    \multirow{2}{*}{Llama-3.1-8B-Instruct}
      & Raw & $74.67$ & $44.93$ & $31.97$ & $5.69$ & $46.27$ \\
      & CorVer & $81.50$ & $53.07$ & $36.69$ & $6.70$ & $49.45$ \\
    \cmidrule(lr){2-7}
    \multirow{2}{*}{Qwen3-8B}
      & Raw & $63.75$ & $30.94$ & $21.45$ & $3.01$ & $42.72$ \\
      & CorVer & $67.39$ & $36.87$ & $27.57$ & $4.92$ & $51.65$ \\
    \cmidrule(lr){2-7}
    \multirow{2}{*}{OLMo-2-13B-Instruct}
      & Raw & $68.62$ & $33.60$ & $26.47$ & $2.20$ & $41.37$ \\
      & CorVer & $81.27$ & $49.00$ & $38.68$ & $5.89$ & $53.73$ \\
    \cmidrule(lr){2-7}
    \multirow{2}{*}{Qwen3-14B}
      & Raw & $65.82$ & $33.19$ & $18.88$ & $2.01$ & $35.86$ \\
      & CorVer & $75.95$ & $44.46$ & $32.02$ & $6.22$ & $53.49$ \\
  \bottomrule
  \end{tabular}
  }
\end{table}

\subsection{PopQA Gains by Popularity and Model Scale}
\label{app:popularity}

Table~\ref{tab:popularity} gives the complete gains underlying Figure~\ref{fig:popularity}, including the full-set gain for each model. The shared Q1--Q4 buckets contain $3{,}573$, $3{,}561$, $3{,}566$, and $3{,}567$ questions, respectively, totaling $14{,}267$. The final row averages the gains equally across the six models.

\begin{table}[H]
  \setlength{\belowcaptionskip}{6pt}
  \centering
  \small
  \setlength{\tabcolsep}{5pt}
  \caption{PopQA accuracy gains (CorVer minus Raw, pp) for six models grouped by scale. Q1 is the least popular quartile and Q4 the most popular. The four buckets contain $3{,}573$, $3{,}561$, $3{,}566$, and $3{,}567$ questions. All covers all $14{,}267$ questions, and Mean gives equal weight to the six models.}
  \label{tab:popularity}
  \resizebox{\textwidth}{!}{%
  \begin{tabular}{llrrrrr}
    \toprule
    Scale & Model & Q1 & Q2 & Q3 & Q4 & All \\
    \midrule
    \multirow{2}{*}{3--4B} & Llama-3.2-3B & $+7.33$ & $+4.66$ & $+6.98$ & $+11.91$ & $+7.72$ \\
     & Qwen3-4B & $+4.79$ & $+3.59$ & $+2.75$ & $+6.59$ & $+4.43$ \\
    \midrule
    \multirow{2}{*}{8B} & Llama-3.1-8B & $+1.54$ & $+3.37$ & $+6.70$ & $+7.26$ & $+4.72$ \\
     & Qwen3-8B & $+7.50$ & $+4.18$ & $+4.04$ & $+8.77$ & $+6.13$ \\
    \midrule
    \multirow{2}{*}{13--14B} & OLMo-2-13B & $+8.59$ & $+7.08$ & $+13.83$ & $+19.37$ & $+12.22$ \\
     & Qwen3-14B & $+12.37$ & $+8.57$ & $+11.44$ & $+20.16$ & $+13.14$ \\
    \midrule
    \multicolumn{2}{l}{Mean across six models} & $+7.02$ & $+5.24$ & $+7.62$ & $+12.34$ & $+8.06$ \\
    \bottomrule
  \end{tabular}
  }
\end{table}

\subsection{Baseline Configuration Comparisons}
\label{app:baseline-comparisons}

FSPO and RLFH configuration comparisons start from the same official base weights with fresh LoRA adapters. They share the $6{,}621$-question pool, rank/scaling $128/256$, seed $42$, and $1{,}000$-token response limit. Each of $100$ steps samples $4$ questions for Qwen3-4B and $3$ for the other models. Final checkpoints use the five-benchmark scoring protocol in Appendix~\ref{app:datasets}.

\subsubsection{FSPO Hyperparameter Comparison}
\label{app:fspo-author-config}

Table~\ref{tab:fspo-author-config} compares FSPO settings from \citet{NEURIPS2025_ddd50f29} (\S6.1 and Appendix~D) with our adapted configuration. The author-key runs use learning rate $4\times10^{-7}$, KL coefficient $0.001$, rollout temperature $1.0$, and $8$ responses per question. The adapted configuration uses $2\times10^{-6}$, $0.04$, $0.7$, and $16$, respectively.

These runs transfer the key hyperparameters to our shared setup. The original FSPO recipe instead uses batches of $8$ questions, a $2{,}048$-token response limit, and one epoch over a mixture of $2$K QA examples and approximately $8$K math problems.

The adapted configuration improves all $20$ model--benchmark scores. Its macro accuracy is higher by $2.72$, $0.36$, $3.74$, and $1.27$ pp on Llama-3.2-3B, Qwen3-4B, Llama-3.1-8B, and Qwen3-8B, respectively. We retain this stronger FSPO configuration in the main comparison.

\begin{table}[htbp]
  \caption{FSPO accuracy ($\%$) with author-reported key hyperparameters (Author key) and our adapted configuration (Main table), evaluated on the same full benchmark sets. Macro is the unweighted mean across benchmarks.}
  \label{tab:fspo-author-config}
  \centering
  \small
  \setlength{\tabcolsep}{4pt}
  \begin{tabular}{@{}lrrrrrr@{}}
    \toprule
    Configuration & TriviaQA & NQ-Open & PopQA & SimpleQA & TruthfulQA & Macro \\
    \midrule
    \multicolumn{7}{l}{\textit{Llama-3.2-3B}} \\
    Author key & 57.96 & 36.81 & 18.55 & 2.50 & 37.82 & 30.73 \\
    Main table & 59.45 & 40.25 & 22.11 & 4.30 & 41.13 & \textbf{33.45} \\
    \midrule
    \multicolumn{7}{l}{\textit{Qwen3-4B}} \\
    Author key & 53.23 & 26.01 & 17.87 & 3.05 & 44.68 & 28.97 \\
    Main table & 53.55 & 26.79 & 18.14 & 3.37 & 44.80 & \textbf{29.33} \\
    \midrule
    \multicolumn{7}{l}{\textit{Llama-3.1-8B}} \\
    Author key & 74.94 & 44.40 & 32.23 & 5.69 & 46.14 & 40.68 \\
    Main table & 79.81 & 51.02 & 36.77 & 6.89 & 47.61 & \textbf{44.42} \\
    \midrule
    \multicolumn{7}{l}{\textit{Qwen3-8B}} \\
    Author key & 64.34 & 31.94 & 21.90 & 3.56 & 44.92 & 33.33 \\
    Main table & 64.77 & 32.24 & 22.77 & 3.91 & 49.33 & \textbf{34.60} \\
    \bottomrule
  \end{tabular}
\end{table}

\FloatBarrier
\subsubsection{RLFH Hyperparameter Comparison}
\label{app:rlfh-author-config}

We also rerun RLFH with author-reported key settings~\citep{wen2025policy}. Both configurations use learning rate $5\times10^{-7}$, $16$ responses per question, and four PPO epochs per batch. Our adapted configuration uses KL coefficient $0.05$ and rollout temperature $0.7$, while the author-key runs use $0.01$ and $1.0$. The author-key runs use top-$p$ $0.9$, top-$k$ $20$, and presence penalty $1.5$. These runs transfer key settings into our training setup rather than reproduce the original recipe in full.

The Qwen judge and reward rules follow the RLFH implementation in Appendix~\ref{app:baselines}.

The preferred configuration varies with the model (Table~\ref{tab:rlfh-author-config}). The author-key settings improve macro accuracy by $6.51$ pp on Llama-3.2-3B and $0.62$ pp on Qwen3-4B. The adapted configuration remains stronger on Llama-3.1-8B and Qwen3-8B by $2.18$ and $1.44$ pp. The main comparison therefore uses the author-key runs for the two smaller models and the adapted runs for the two $8$B models. Each selected row comes from one checkpoint across all five benchmarks.

\begin{table}[htbp]
  \caption{RLFH accuracy ($\%$) with our adapted configuration and author-reported key settings. Bold macro values identify the complete rows selected for Table~\ref{tab:main}.}
  \label{tab:rlfh-author-config}
  \centering
  \small
  \setlength{\tabcolsep}{4pt}
  \begin{tabular}{@{}lrrrrrr@{}}
    \toprule
    Configuration & TriviaQA & NQ-Open & PopQA & SimpleQA & TruthfulQA & Macro \\
    \midrule
    \multicolumn{7}{l}{\textit{Llama-3.2-3B}} \\
    Adapted & 51.98 & 26.95 & 17.64 & 2.15 & 25.34 & 24.81 \\
    Author key & 59.12 & 36.90 & 20.75 & 2.61 & 37.21 & \textbf{31.32} \\
    \midrule
    \multicolumn{7}{l}{\textit{Qwen3-4B}} \\
    Adapted & 50.18 & 25.01 & 16.22 & 1.78 & 33.17 & 25.27 \\
    Author key & 50.68 & 24.60 & 15.99 & 1.60 & 36.60 & \textbf{25.89} \\
    \midrule
    \multicolumn{7}{l}{\textit{Llama-3.1-8B}} \\
    Adapted & 72.49 & 41.22 & 29.92 & 5.36 & 38.92 & \textbf{37.58} \\
    Author key & 71.62 & 38.42 & 29.03 & 4.37 & 33.54 & 35.40 \\
    \midrule
    \multicolumn{7}{l}{\textit{Qwen3-8B}} \\
    Adapted & 62.77 & 30.39 & 20.52 & 2.91 & 42.35 & \textbf{31.79} \\
    Author key & 62.20 & 28.86 & 19.82 & 2.94 & 37.94 & 30.35 \\
    \bottomrule
  \end{tabular}
\end{table}

\FloatBarrier
\subsubsection{FoRAG Batch Configuration Comparison}
\label{app:forag-recovery}
FoRAG releases model weights and datasets but not training code. The original paper specifies SFT settings, while its RL formulation leaves the numeric PPO learning rate, KL coefficient, and rollout temperature unspecified~\citep{cai2024forag}. Table~\ref{tab:baseline-config} reports the settings used in our training setup.

\paragraph{Rollout allocation.}
We compare the initial and adjusted rollout allocations in Table~\ref{tab:rollout-config} on all four main models. Both use LoRA rank/scaling $128/256$, learning rate $5\times10^{-7}$, KL coefficient $0.02$, and four PPO epochs per batch. The main table uses adjusted runs for Llama-3.2-3B and Qwen3-8B and initial runs for the other models, all at rollout temperature $1.0$. On Qwen3-8B, reducing generation from $4{,}800$ to $400$ responses improves all five accuracies (Table~\ref{tab:forag-recovery}).

\begin{table}[htbp]
  \caption{Qwen3-8B FoRAG accuracy under two rollout allocations. Accuracy is evaluated on all $40{,}964$ questions. Response counts describe training generation budgets. The adjusted run takes $5$ h $42$ min on H100 80GB.}
  \label{tab:forag-recovery}
  \centering
  \small
  \setlength{\tabcolsep}{5pt}
  \begin{tabular}{lrrrrrr}
    \toprule
    Training responses & TriviaQA & NQ-Open & PopQA & SimpleQA & TruthfulQA & Macro \\
    \midrule
    $4{,}800$ (original) & 49.14 & 19.11 & 15.11 & 1.48 & 21.18 & 21.20 \\
    $400$ (adjusted) & 60.86 & 28.42 & 19.81 & 2.73 & 41.00 & 30.57 \\
    \bottomrule
  \end{tabular}
\end{table}

\FloatBarrier
\subsubsection{KnowRL Training Duration Comparison}
\label{app:knowrl-training}
We compare step-$100$ and step-$200$ checkpoints on all four main models using the KnowRL configuration in Table~\ref{tab:baseline-config}. Each run continues for $100$ additional optimizer steps with hardware, base-model revision, and training pool held fixed. Main accuracy and cost comparisons use step $100$ and the timing scope in Appendix~\ref{app:cost}. The duration diagnostics below focus on Qwen3-4B and Qwen3-8B at step $200$, with complete training times of $21$ h $54$ min and $36$ h $23$ min, respectively.

\paragraph{Reward configuration.}
In our KnowRL configuration~\citep{ren_knowrl_2026}, the correctness term assigns $+2$ to correct answers, $+1$ to refusals, and $-1$ to incorrect answers. CorVer assigns $+2$ to correct answers and $-1$ to incorrect, refused, or empty answers (Appendix~\ref{app:reward-config}). In answer-accuracy evaluation, refused and empty answers receive no credit and remain in the denominator for every method.

\paragraph{Accuracy and refusal.}
\label{app:knowrl-diagnostic}
The original KnowRL study~\citep{ren_knowrl_2026} examines $100$--$300$ training steps and reports incorrect rate, refusal rate, precision on answered questions, and F1. Extending both Qwen runs from $100$ to $200$ optimizer steps did not improve macro accuracy. Qwen3-8B macro accuracy falls from $30.48\%$ to $28.58\%$ under the same evaluation protocol. For Qwen3-4B, macro accuracy falls from $20.80\%$ to $19.07\%$, with refusal increasing on all five benchmarks.

The Qwen3-4B step-$200$ outputs contain $26{,}868$ explicit refusals among $40{,}964$ responses. The high refusal rate is consistent with KnowRL's abstention incentive and limits accuracy when refusals receive no credit. Table~2 of \citet{ren_knowrl_2026} reports a related trade-off on Skywork-OR1-7B-Preview: the SimpleQA refusal rate rises from $21.33\%$ to $37.67\%$ while the incorrect rate falls from $76.33\%$ to $60.33\%$, with precision on answered questions moving from $2.97$ to $3.21$ and F1 from $2.61$ to $2.46$. ChineseSimpleQA shows the same pattern ($24.00\%$ to $40.33\%$ refusal, $67.00\%$ to $52.33\%$ incorrect). These results align with KnowRL's goal of avoiding incorrect answers through appropriate abstention. Its training-step analysis likewise reports refusal rates increasing and then plateauing over $100$--$300$ steps~\citep[\S4.5]{ren_knowrl_2026}. Precision on answered questions includes only attempted answers, whereas our accuracy retains all questions and scores refusals as errors. The observed accuracy loss therefore reflects the cost of abstention under an answer-required metric, while reliability among attempted answers is a separate question.

\subsection{Sensitivity to Answer Scoring}
\label{app:scoring-checks}

\paragraph{One-way substring matching.}
We rescore the Llama-3.2-3B-Instruct answers on all $35{,}821$ TriviaQA, NQ-Open, and PopQA questions after removing the reverse-match branch. A match now requires the final answer to contain a reference answer or alias, with the remaining normalization and scoring rules unchanged. Table~\ref{tab:substring-check} shows that CorVer retains its lead. Its macro accuracy falls by $0.31$ pp, compared with $1.48$ for Raw and $1.24$ for FSPO. The advantage over FSPO increases from $5.38$ to $6.31$ pp under the one-way rule.

\begin{table}[htbp]
  \caption{Full-set substring check for Llama-3.2-3B-Instruct. Reverse-only counts are pooled across the three datasets. Accuracy ($\%$) is the unweighted mean of their individual accuracies.}
  \label{tab:substring-check}
  \centering
  \small
  \setlength{\tabcolsep}{8pt}
  \begin{tabular}{@{}lrrr@{}}
    \toprule
    Method & Reverse-only matches & Bidirectional & One-way \\
    \midrule
    Raw & 457 & 37.68 & 36.20 \\
    FSPO & 432 & 40.60 & 39.36 \\
    CorVer & 132 & \textbf{45.98} & \textbf{45.67} \\
    \bottomrule
  \end{tabular}
\end{table}

\paragraph{Semantic correctness on three QA datasets.}
We also score $100$ questions from each of TriviaQA, NQ-Open, and PopQA using a remote GLM judge. Raw, FSPO, and CorVer are evaluated on the same $300$ questions, giving $900$ answers. The same judge and grading rules apply to all answers. Table~\ref{tab:qa-semantic-check} compares the original substring scores with semantic correctness on this same sample. CorVer leads under both measures, reaching $54.33\%$ semantic correctness versus $47.33\%$ for FSPO and $44.00\%$ for Raw.

\begin{table}[htbp]
  \caption{Paired answer-scoring comparison ($\%$) for Llama-3.2-3B-Instruct on $300$ questions, with $100$ from each of three QA datasets.}
  \label{tab:qa-semantic-check}
  \centering
  \small
  \setlength{\tabcolsep}{12pt}
  \begin{tabular}{@{}lrr@{}}
    \toprule
    Method & Original substring & Semantic correctness \\
    \midrule
    Raw & 42.00 & 44.00 \\
    FSPO & 47.00 & 47.33 \\
    CorVer & \textbf{52.33} & \textbf{54.33} \\
    \bottomrule
  \end{tabular}
\end{table}

\paragraph{Truthfulness and informativeness on TruthfulQA.}
Using the same GLM judge, we assess the three methods on a shared sample of $100$ TruthfulQA questions. We report the proportion of answers judged truthful and the proportion judged both truthful and informative, each over all $100$ questions. CorVer reaches $55\%$ truthfulness and $54\%$ truthful-and-informative answers (Table~\ref{tab:truthfulqa-semantic-check}). Its advantage on this sample therefore extends to semantic judgments beyond the main-table ROUGE rule. These sample-based checks supplement the full-set lexical scores.

\begin{table}[htbp]
  \caption{GLM semantic evaluation ($\%$) of Llama-3.2-3B-Instruct on the same $100$ TruthfulQA questions across methods. The second metric requires both truthfulness and informativeness.}
  \label{tab:truthfulqa-semantic-check}
  \centering
  \small
  \setlength{\tabcolsep}{12pt}
  \begin{tabular}{@{}lrr@{}}
    \toprule
    Method & Truthful & Truthful and informative \\
    \midrule
    Raw & 44 & 34 \\
    FSPO & 46 & 40 \\
    CorVer & \textbf{55} & \textbf{54} \\
    \bottomrule
  \end{tabular}
\end{table}

\subsection{Accuracy and Training Cost on H100}
\label{app:cost}

\paragraph{Measurement and comparison scope.}
Figure~\ref{fig:feasibility}(b) and Table~\ref{tab:cross-method-cost} pair accuracy and complete training time at step $100$ for the selected Qwen3-4B and Qwen3-8B configurations. Each policy trains on one H100 80GB. Timing covers initialization, rollout generation, reward computation, policy updates, and final saving, excluding job queueing and post-training evaluation. RLFH and KnowRL use remote judges: their inference, API communication, and response waiting contribute to elapsed time, while remote computation occurs outside the local H100. FSPO and FoRAG score locally. Raw provides an accuracy reference.

The selected RLFH runs use author-key settings on Qwen3-4B and adapted settings on Qwen3-8B. FoRAG uses the initial Qwen3-4B allocation and adjusted Qwen3-8B allocation (Appendix~\ref{app:baseline-comparisons}). Workloads differ in generation budgets and update schedules: PPO reuses responses for four epochs, and adjusted Qwen3-8B FoRAG generates $400$ responses versus CorVer's $4{,}800$. Time ratios therefore compare complete training configurations rather than isolated verifier speed at matched workload.

\paragraph{Additional CorVer timings.}
Under the same timing scope, $100$-step CorVer training takes $1$ h $50$ min on Qwen3-14B and $2$ h $17$ min on OLMo-2-13B-Instruct. These models are excluded from the two-model averages.

\paragraph{Aggregation.}
Accuracy averages the five benchmarks equally, and accuracy and time averages weight the two models equally. Each time ratio divides a baseline's mean complete-run duration by CorVer's. Calculations use unrounded values, while the table displays h:mm rounded to the nearest minute.

\begin{table*}[!htbp]
  \caption{Accuracy and training cost. All timed runs use one H100 80GB. All methods are compared at step $100$, using the same accuracy evaluations as Table~\ref{tab:main}. Macro is five-benchmark macro accuracy (\%). Time is training duration in h:mm (rounded to the nearest minute), excluding queueing and evaluation. Mean weights the two models equally. The last column gives mean time relative to CorVer.}
  \label{tab:cross-method-cost}
  \centering
  \small
  \setlength{\tabcolsep}{4pt}
\begin{tabular}{lrrrrrrrr}
\toprule
& & \multicolumn{2}{c}{Qwen3-4B} & \multicolumn{2}{c}{Qwen3-8B} & \multicolumn{2}{c}{Mean} & Time ratio \\
\cmidrule(lr){3-4}\cmidrule(lr){5-6}\cmidrule(lr){7-8}
Method & Steps & Macro & Time & Macro & Time & Macro & Time & / CorVer \\
\midrule
Raw & \textemdash & 28.62 & \textemdash & 32.37 & \textemdash & 30.50 & \textemdash & \textemdash \\
\textbf{CorVer} & \textbf{100} & \textbf{33.22} & \textbf{2:01} & \textbf{37.68} & \textbf{1:36} & \textbf{35.45} & \textbf{1:49} & 1.00$\times$ \\
FSPO & 100 & 29.33 & 10:58 & 34.60 & 9:04 & 31.97 & 10:01 & 5.52$\times$ \\
RLFH & 100 & 25.89 & 15:30 & 31.79 & 22:15 & 28.84 & 18:53 & 10.41$\times$ \\
FoRAG & 100 & 26.96 & 22:28 & 30.57 & 5:42 & 28.76 & 14:05 & 7.76$\times$ \\
KnowRL & 100 & 20.80 & 12:58 & 30.48 & 18:51 & 25.64 & 15:55 & 8.77$\times$ \\
\bottomrule
\end{tabular}
\space
\end{table*}


\end{document}